\documentclass[manuscript]{acmart}
\renewcommand\footnotetextcopyrightpermission[1]{} 

\AtBeginDocument{%
  \providecommand\BibTeX{{%
    \normalfont B\kern-0.5em{\scshape i\kern-0.25em b}\kern-0.8em\TeX}}}

\usepackage[utf8]{inputenc}
\usepackage{graphics}
\usepackage{graphicx}
\usepackage{amsmath,amsfonts}
\usepackage{algorithm}
\usepackage{algorithmic}
\usepackage{multirow}
\usepackage{afterpage}
\usepackage{url}
\usepackage{float}

\newcommand{\softmax}{\ensuremath{\mathrm{softmax}}}
\DeclareMathOperator*{\argmax}{arg\,max}

\begin{document}

\fancyfoot{}
\title{Topological Effects on Attacks Against Vertex Classification}
\titlenote{This research was sponsored in part by the Combat Capabilities Development Command Army Research Laboratory and was accomplished under Cooperative Agreement Number W911NF-13-2-0045 (ARL Cyber Security CRA) and by the United States Air Force under Air Force Contract No. FA8702-15-D-0001. The views and conclusions contained in this document are those of the authors and should not be interpreted as representing the official policies, either expressed or implied, of the Combat Capabilities Development Command Army Research Laboratory, the United States Air Force, or the U.S. Government. The U.S. Government is authorized to reproduce and distribute reprints for Government purposes not withstanding any copyright notation here on.}
\author{Benjamin A. Miller}
\email{miller.be@northeastern.edu}
\affiliation{%
  \institution{Northeastern University}
  \city{Boston}
  \state{MA}
  \postcode{02115}
}
\author{Mustafa \c{C}amurcu}
\email{camurcu.m@northeastern.edu}
\affiliation{%
  \institution{Northeastern University}
  \city{Boston}
  \state{MA}
  \postcode{02115}
}
\author{Alexander J. Gomez}
\email{gomez.alexa@northeastern.edu}
\affiliation{%
  \institution{Northeastern University}
  \city{Boston}
  \state{MA}
  \postcode{02115}
}
\author{Kevin Chan}
\email{kevin.s.chan.civ@mail.mil}
\affiliation{%
  \institution{Army Research Laboratory}
  \city{Adelphi}
  \state{MD}
  \postcode{20783}
}
\author{Tina Eliassi-Rad}
\email{t.eliassirad@northeastern.edu}
\affiliation{%
  \institution{Northeastern University}
  \city{Boston}
  \state{MA}
  \postcode{02115}
}

\renewcommand{\shortauthors}{Miller et al.}

\begin{abstract}
    Vertex classification is vulnerable to perturbations of both graph topology and vertex attributes, as shown in recent research. As in other machine learning domains, concerns about robustness to adversarial manipulation can prevent potential users from adopting proposed methods when the consequence of action is very high. This paper considers two topological characteristics of graphs and explores the way these features affect the amount the adversary must perturb the graph in order to be successful. We show that, if certain vertices are included in the training set, it is possible to substantially an adversary's required perturbation budget. On four citation datasets, we demonstrate that if the training set includes high degree vertices or vertices that ensure all unlabeled nodes have neighbors in the training set, we show that the adversary's budget often increases by a substantial factor---often a factor of 2 or more---over random training for the Nettack poisoning attack. Even for especially easy targets (those that are misclassified after just one or two perturbations), the degradation of performance is much slower, assigning much lower probabilities to the incorrect classes. In addition, we demonstrate that this robustness either persists when recently proposed defenses are applied, or is competitive with the resulting performance improvement for the defender.
\end{abstract}

\maketitle
\thispagestyle{empty}

\section{Introduction}
\label{sec:intro}
Classification of vertices in graphs is an important problem in a variety of applications, from e-commerce (classifying users for targeted advertising) to security (classifying computer nodes as malicious or not) to bioinformatics (classifying roles in a protein interaction network). In the past several years, numerous methods have been developed for this task (see, e.g.,~\cite{graphsage, Moore2017}). More recently, research has focused on attacks by adversaries~\cite{Zugner2018, Dai2018} and robustness to such attacks~\cite{Wu2019}. If an adversary were able to insert misleading data into the training set (e.g., generate benign traffic during a data collection period that could cover its behavior during testing/inference time), the chance of successfully evading detection would increase (which is undesirable to the data analyst).

To classify vertices in the presence of adversarial activity, we must implement learning systems that are robust to such potential manipulation. If such malicious behavior has low cost to the attacker and imposes high cost on the data analyst, machine learning systems will not be trusted and adopted for use in practice, especially in high-stakes scenarios such as network security and traffic safety. Understanding how to achieve robustness is key to realizing the full potential of machine learning.

Adversaries, of course, will attempt to conceal their manipulation. In a recent paper, Z\"{u}gner et al.~propose an adversarial technique called Nettack~\cite{Zugner2018}, which can create perturbations that are subtle while still being extremely effective in decreasing performance on the target vertices. The authors use their poisoning attack against a graph convolutional network (GCN). 

From a defender's perspective, we aim to make it more difficult for the attacker to cause node misclassification. In addition to changing the properties of the classifier itself, there may be portions of a complex network that provide more information for learning than others. Complex networks are highly heterogeneous and random sampling may not be the best way to obtain labels. If there is flexibility in the means of obtaining training data, the defender should leverage what is known about the graph topology.

This paper demonstrates that leveraging complex network properties can improve robustness of GCNs in the presence of adversaries. We focus on two alternative techniques for training data selection. In both methods, we aim to train with a subset of nodes that are well connected to the held out set. Here we see a benefit, often raising the number of perturbations required for a given level of attack success by a factor of 2 to 4. When it is possible to pick a specific subset on which to train, this can provide a significant advantage. Some combination of these methods will likely be useful to develop a more robust vertex classification system.

The contributions of this work are as follows:
\begin{itemize}
    \item We propose two methods---\textsc{StratDegree} and \textsc{GreedyCov\-er}---for selecting training data that put a greater burden on attackers.
    \item We demonstrate that the robustness gained via these methods cannot be reliably obtained by simply increasing the amount of randomly selected training data.
    \item We adapt Nettack so that the attacker is aware of the training set selection method and illustrate that the advantage is maintained when the attacker adapts.
    \item We show that the data selection methods are competitive with defenses recently proposed in the literature.
    \item We identify a potential tradeoff between overall classifier performance and robustness to attack.
\end{itemize}
These contributions all point toward interesting future research in this area, such as determining the conditions under which such methods are effective.\footnote{An early version of this work appeared in~\cite{Miller2019}.}

The remainder of this paper is organized as follows. In Section~\ref{sec:model} we describe the vertex classification problem and the Nettack method. Section~\ref{sec:methods} details the methods we investigate to provide greater robustness, and Section~\ref{sec:setup} outlines the experimental setup, including adaptations of Nettack and descriptions of defenses from the open literature. Section~\ref{sec:results} documents several experimental results, illustrating the effectiveness of the proposed methods. In Section~\ref{sec:related}, we briefly contextualize our work within the current literature. In Section~\ref{sec:conclusion} we conclude with a summary and outline open problems and future work.

\section{Problem Model}
\label{sec:model}
We consider the vertex classification problem as described in~\cite{Zugner2018}, where we are given a graph $G=(V,E)$ of size $N=|V|$ and an $N\times d$ matrix of vertex attributes $X$. Each node has an arbitrary numeric index from 1 to $N$. For this work, as in~\cite{Zugner2018}, we consider only binary attributes. In addition to its $d$ attributes, each node has a label denoting its class. We enumerate classes as integers from $1$ to $C$. Given a subset of labeled instances, the goal is to correctly classify the unlabeled nodes.

The focus of~\cite{Zugner2018} is on GCNs, which make use of the adjacency matrix for the graph $A=\{a_{ij}\}$, where $a_{ij}$ is 1 if there is an edge between node $i$ and node $j$ and is 0 otherwise. The GCN applies a symmetrized graph convolution to the input layer. That is, if we let $D$ be the diagonal matrix of vertex degrees---i.e., the $i$th diagonal entry is the number of edges connected to vertex $i$, $d_{ii}=\sum_{j=1}^N{a_{ij}}$---then the output of the first layer of the network is expressed as
\begin{equation*}
    H = \sigma\left(D^{-1/2}AD^{-1/2}XW_1\right),
\end{equation*}
where $W_1$ is a weight matrix, $X$ is a feature matrix whose $i$th row is $x_i^T$ (the attribute vector for row vertex $i$), and $\sigma$ is the rectifier function. From the hidden layer to the output layer, a similar graph convolution is performed, followed by a softmax output:
\begin{equation*}
    Y = \softmax\left(D^{-1/2}AD^{-1/2}HW_2\right).\
\end{equation*}
The focus in~\cite{Zugner2018} is on GCNs with a single hidden layer. Each vertex is then classified according to the largest entry in the corresponding row of $Y$.

The vertex attack proposed in~\cite{Zugner2018} operates on a surrogate model where the rectifier function is replaced by a linear function, thus approximating the overall network as
\begin{align}
    Y&\approx \softmax\left(\left(D^{-1/2}AD^{-1/2}\right)^2XW_1W_2\right)\label{eq:surrogate}\\
    &=\softmax\left(\left(D^{-1/2}AD^{-1/2}\right)^2XW\right).\nonumber
\end{align}
Nettack uses a greedy algorithm to determine how to perturb both $A$ and $X$ to make the GCN misclassify a target node. The changes are intended to be ``unnoticeable,'' i.e., the degree distribution of $G$ and the co-occurrence of features are changed negligibly. Using the approximation in (\ref{eq:surrogate}), Nettack perturbs by either adding or removing edges or turning off binary features so that the classification margin is reduced the most at each step. Note that while it can change the topology and the features, Nettack does \emph{not} change the labels of any vertices. An additional variation on Nettack allows either ``direct'' attacks, in which the target node itself has its edges and features changed, or indirect ``influencer'' attacks, where the neighbors of the target have their data altered.

The classifier is evaluated in a context where only some of the labels are known, and the labeled data are split into training and validation sets. 
To train the GCN, 10\% of the data are selected at random (or by one of the alternative methods outlined in Section~\ref{sec:methods}), and another 10\% is selected for validation. The remaining 80\% is the test data. After training, nodes are selected for attack among those that are correctly classified: the 10 where the margin is largest, the 10 where the margin is smallest, and 20 more at random. Each of these is taken as a target node for attack in an experiment. The attack is evaluated based on how much the classification margin decreases for the targeted nodes.

Taking the perspective of a defender against the attack, we want to ensure that this decrease in margin is minimized. The goal is to determine how to make the classifier as robust as possible to this attack, making the attacker as ineffective as possible for any given level of perturbation.

\section{Proposed Methods}
\label{sec:methods}

As we investigated classification performance using Nettack, we noted that nodes in the test set with many neighbors in the training set were more likely to be correctly classified. This dependence on labeled neighbors is consistent with previous observations~\cite{Neville2009}. We observed this effect using the standard method of training data selection used in the original Nettack paper: randomly select 10\% for training, 10\% for validation, and 80\% for testing. This observation suggested that a training set that provides something like a vertex cover---a kind of ``scaffolding'' for the unlabeled data---could make the classification more robust.


We considered two methods to test this hypothesis. The first simply chooses the highest-degree nodes to be in the training set. The top 10\% of each class is chosen for the training data, and the remaining 90\% is randomly split (stratified by class) into test (80\%) and validation (10\%), maintaining the proportions of the original experiment in~\cite{Zugner2018}. We refer to the stratified degree-based thresholding method as \textsc{StratDegree}. The other method uses a greedy approach in an attempt to ensure every node has at least a minimal number of neighbors in the training set. Starting with an empty training set and a threshold $k=0$, we iteratively add the node with the largest number of neighbors connected to at most $k$ nodes in the training set. When there are no such neighbors, we increment $k$. This procedure continues until we have the desired proportion of the overall dataset for training (again, 10\% in our experiments). The remaining data are randomly partitioned into test and validation sets. Algorithm~\ref{alg:cover} provides the pseudo-code. In this case, there is no stratification by class. Incorporating this aspect is part of our future work.

\begin{algorithm}
\begin{centering}
\begin{algorithmic}
\STATE \textbf{Input:} Graph $G=(V,E)$, training proportion $t\in (0,1)$
\STATE \textbf{Output:} Training set $T\subset V$
\STATE $k\gets 0$, $T\gets \emptyset$
\FORALL{$u\in V$}
\STATE $m_u\gets 0$ \ \ \ \ $\langle\langle$mark all nodes 0$\rangle\rangle$
\ENDFOR
\WHILE{$|T|<|V|t$}
    \STATE $v \gets \argmax_{u\in V\setminus T} \sum_{u^\prime\in\mathcal{N}(u)}\mathbb{I}\left[m_{u^\prime}=k\right]$
    \IF{$\sum_{u^\prime\in\mathcal{N}(v)}\mathbb{I}\left[m_{u^\prime}=k\right]=0$}
        \STATE {$k\gets k+1$} \ \ $\langle\langle$incr. min. num. trained neighbors$\rangle\rangle$
    \ELSE
        \STATE{$T\gets T\cup \{v\}$}
        \STATE $m_v\gets -1$
        \FORALL {$u^\prime\in \mathcal{N}(v)\setminus T$}
            \STATE $m_{u^\prime}\gets m_{u^\prime}+1$
        \ENDFOR
    \ENDIF
\ENDWHILE
\RETURN $T$
\end{algorithmic}
\caption{\textsc{GreedyCover}}
\label{alg:cover}
\end{centering}
\end{algorithm}

Both of these approaches raise an issue. Since Nettack adds edges to the graph, it changes the degree of the nodes. It may also change which nodes are the best at covering the unlabeled data. The targets, however, are selected from among the test set (the nodes with ``unknown'' labels). These two aspects of the problem setup are in contention: selecting the target requires knowledge of the training set, but the training set selection requires knowledge of degree, which is not fully known until after the attack. In Section~\ref{subsec:adaptations}, we propose a relatively conservative adaptation of Nettack that avoids changing the training set. A deeper investigation of the interplay between attacks and defenses will be a focus of future work.

\subsection{Computational Complexity}
Using \textsc{StratDegree} and \textsc{GreedyCover} both have computational costs beyond random sampling. \textsc{StratDegree} requires finding the highest-degree nodes, which, for a constant fraction of the dataset size, will require $O(|E|+|V|\log{|V|})$ time (for computing degrees and sorting), compared to linear time for random sampling. Each step in \textsc{GreedyCover} requires finding the vertex with the most neighbors minimally connected to the training set. As written in Algorithm~\ref{alg:cover}, each iteration requires $O(|E|)$ time to count the number of such neighbors each node has, which would result in an overall running time of $O(|V||E|)$. This could be improved using a priority queue---such as a Fibonacci heap---to achieve $O(|E|+|V|\log{|V|})$ time ($O(|V|)$ logarithmic-time min. extractions and $O(|E|)$ constant-time key updates). Thus, the two proposed method require moderate overhead compared to the running time for the GCN.
\section{Experimental Setup}
\label{sec:setup}
In each experiment, we choose a graph, randomly select a test/val\-i\-da\-tion split (after either deterministically or randomly selecting the training set), select 40 targets as discussed in Section~\ref{sec:model}, and perturb the graph either directly or indirectly to change the target's classification. We evaluate the perturbation by inspecting the classification margin for the target node. Specifically, the margins reported here are the log probability ratios for the correct class versus the highest probability incorrect class, i.e., if the true class is $c$, the margin is $$\log\frac{p_c}{\max_{c^\prime\neq c}{p_{c^\prime}}}.$$ Thus, if the vertex is correctly classified, the margin will be positive; otherwise it will be negative.

For each target, we perturb up to 50 times. We do the following to determine the budget required for an attacker to achieve a given probability of success. First, compute the associated quantile of the distribution of margins across targets (e.g., 10th percentile for 10\% attacker success probability) as a function of number of perturbations. We then determine where this function first becomes nonpositive (i.e., the smallest number of perturbations for the attacker to successfully cause the target to be misclassified). This value is the required budget. We linearly interpolate between values for better resolution. If a given level of attack success is never achieved within the perturbations attempted, we set the required budget to number of perturbations plus 1. We average this value over 5 trials and report standard errors in the budget plots in Section~\ref{sec:results}.

\subsection{Datasets}
We use the three datasets used in the Nettack paper in our experiments, plus one larger citation dataset:
\begin{itemize}
    \item \textbf{CiteSeer} The CiteSeer dataset has 3312 scientific publications put into 6 classes. The network has 4732 links representing citations between the publications. The features of the nodes contain 1s and 0s indicating the presence of the word in the paper. There are 3703 unique words considered for the dictionary.
    
    \item \textbf{Cora} The Cora dataset consists of 2708 machine learning papers classified into one of seven categories. The citation network consists of 5429 citations. For each paper (vertex) in the network there is a feature vector of 0s and 1s for whether it contains one of 1433 unique words.
    
    \item \textbf{PolBlogs} The political blogs dataset consists of 1490 blogs labeled as either liberal or conservative. A total of 19,025 links between blogs form the directed edges of the graph. No attributes are used.
    
    \item \textbf{PubMed} The PubMed dataset consists of 19,717 papers pertaining to diabetes classified into one of three classes. The citation network consists of 44,338 citations. For each paper in the network there is a binary feature vector representing the presence of 500 words.
\end{itemize}

\subsection{Defenses and Adaptations}
\label{subsec:adaptations}
\subsubsection{Number of Trained Neighbors with Random Selection}
In addition to testing robustness of the proposed data selection methods, we are interested in whether the same level of robustness can be obtained with a larger amount of training data. Thus, in addition to randomly selecting 10\% of the data for training, we randomly select up to 30\% of the data in 5\% increments. In all cases 10\% of the data are used for validation. At each point, in addition to evaluating the robustness of the classifier based on the required adversary budget, we measure the average number of neighbors connected to a node outside of the training set, i.e., for the training set $T\subset V$, we record
\begin{equation}
    \frac{1}{\left|V\setminus T\right|}\sum_{i\in T}\sum_{j\in V\setminus T}{a_{ij}}.\label{eq:avgLabeledNeighbors}
\end{equation} 
This will allow us to evaluate what impact the overall number of connections to the training data has on performance, and whether performance with the proposed training data selection methods match any trend observed with random training.

\subsubsection{Adapted Nettack}
As mentioned in Section~\ref{sec:methods}, there is an inherent tension between Nettack, which changes the graph topology, and the training set selection methods, which use topological properties to select training nodes. As a first-order adaptation of Nettack to these methods, we altered Nettack so that when it considers proposed perturbations, it does not make perturbations that would change the training dataset. When selecting based on degree, this involves filtering any perturbations that would push a node across the degree threshold, making a non-training node's degree high enough to be a selected or sufficiently reducing a training node's degree. When using \textsc{GreedyCover}, we filter perturbations that could increase (or decrease) the number of neighbors that are minimally connected to the training set to move a node into (or out of) the training set. In both cases we assume the attacker knows the selection method in use. The code to modify Nettack for these adaptations is provided in the supplement.

\subsubsection{Proposed Defenses}
Defenses have been proposed since Nettack was published. Here, we consider two such defenses: removing edges between nodes with no common attributes as in~\cite{Wu2019}, and using a low-rank approximation for the adjacency matrix as  in~\cite{Entezari2020}. In the low-rank case, we use the first 10 components of the singular value decomposition.

\section{Empirical Results}
\label{sec:results}
\subsection{Impact of Training Methods}
\label{subsec:varyTrain}
\begin{figure*}
    \centering
    \includegraphics[width=1.96in]{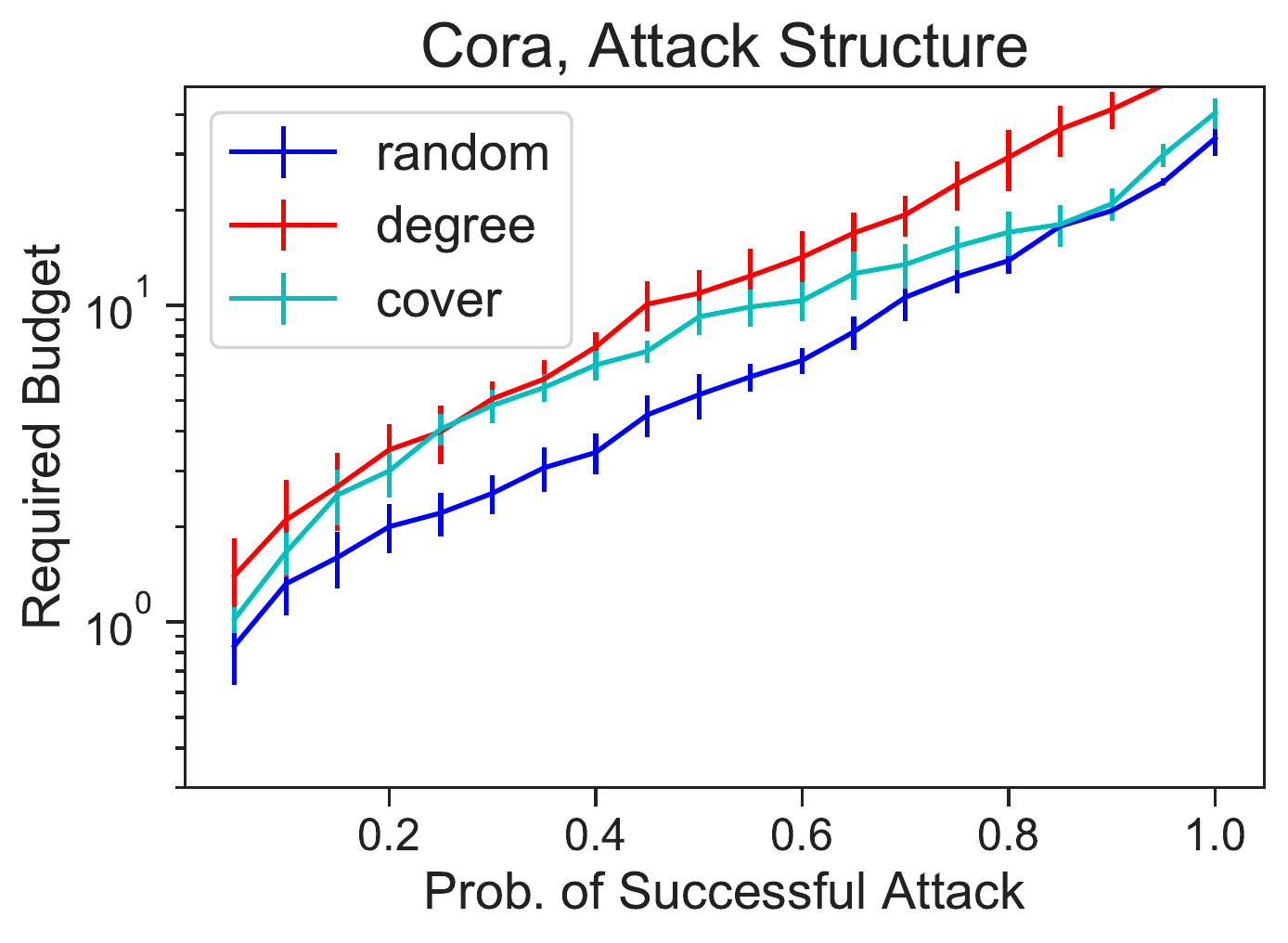}
    \includegraphics[width=1.96in]{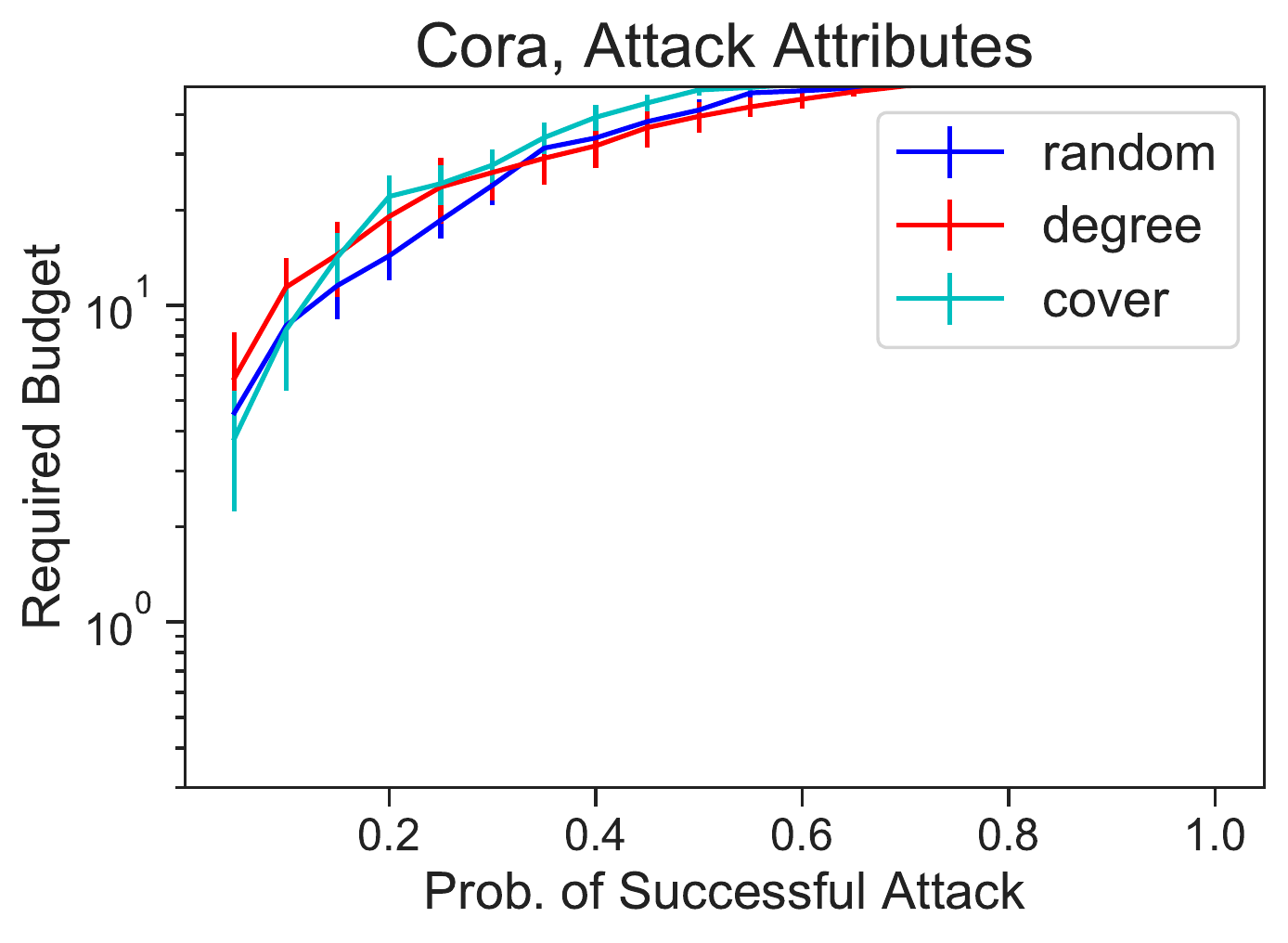}
    \includegraphics[width=1.96in]{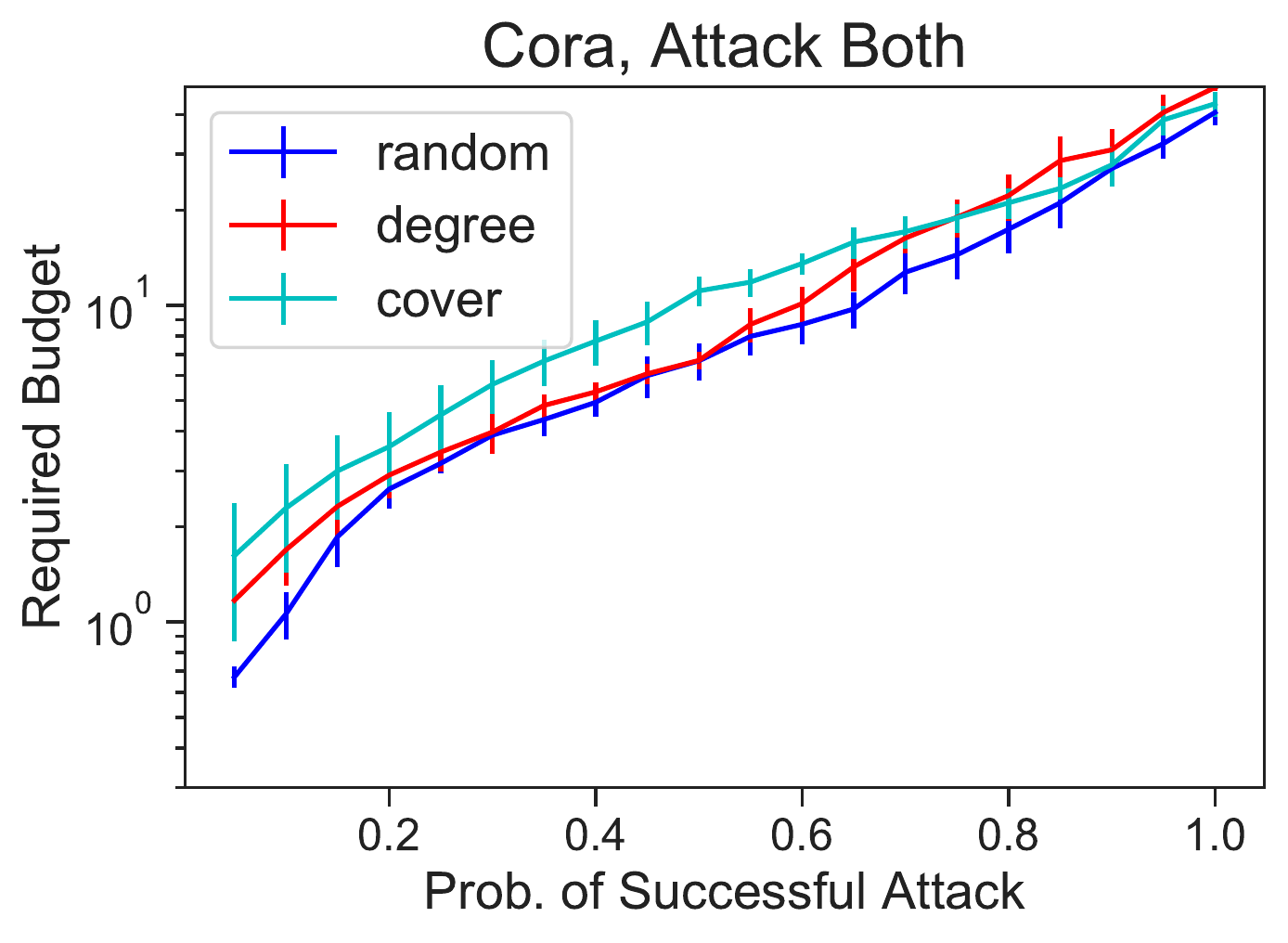}\\
    \includegraphics[width=1.96in]{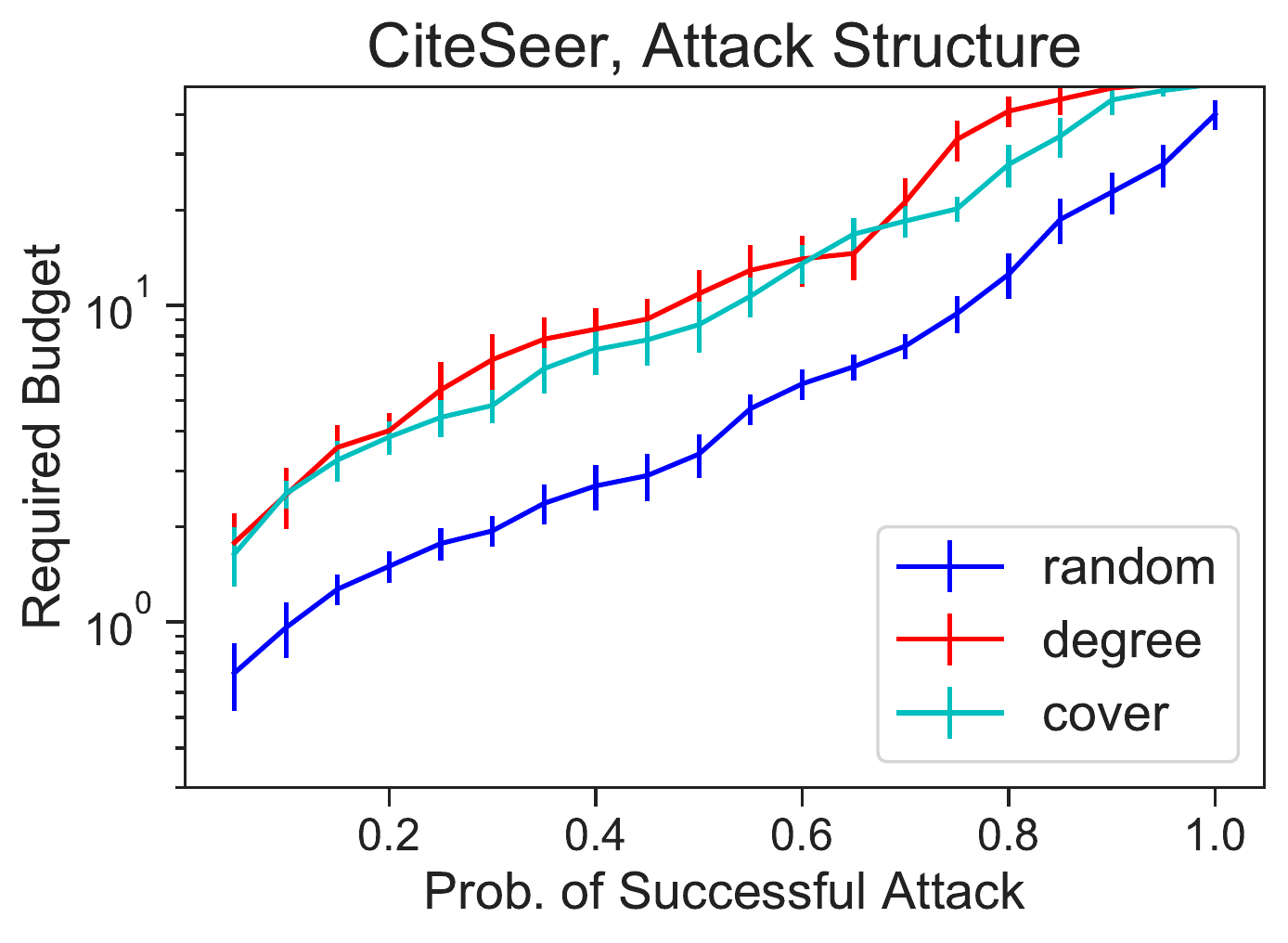}
    \includegraphics[width=1.96in]{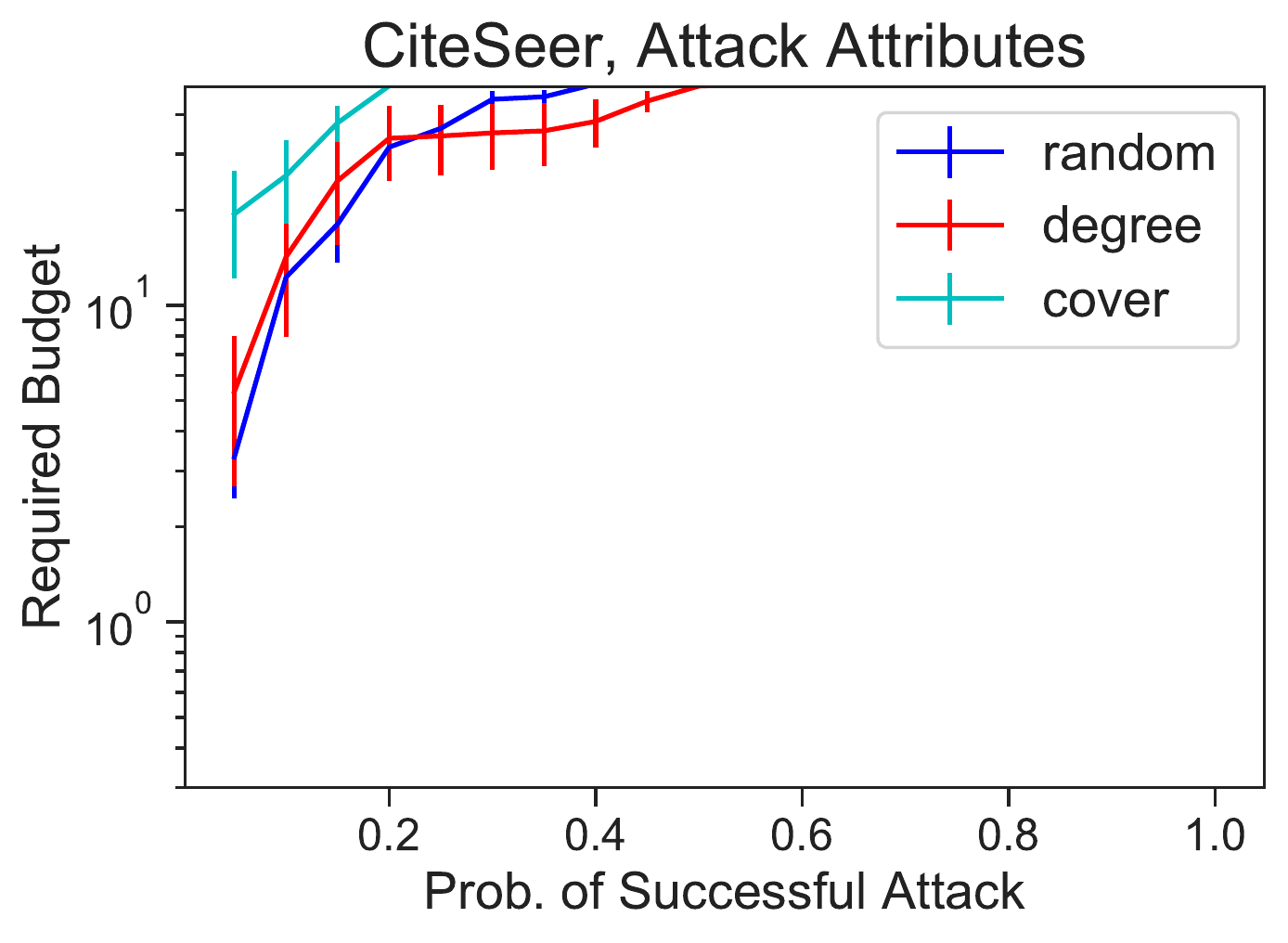}
    \includegraphics[width=1.96in]{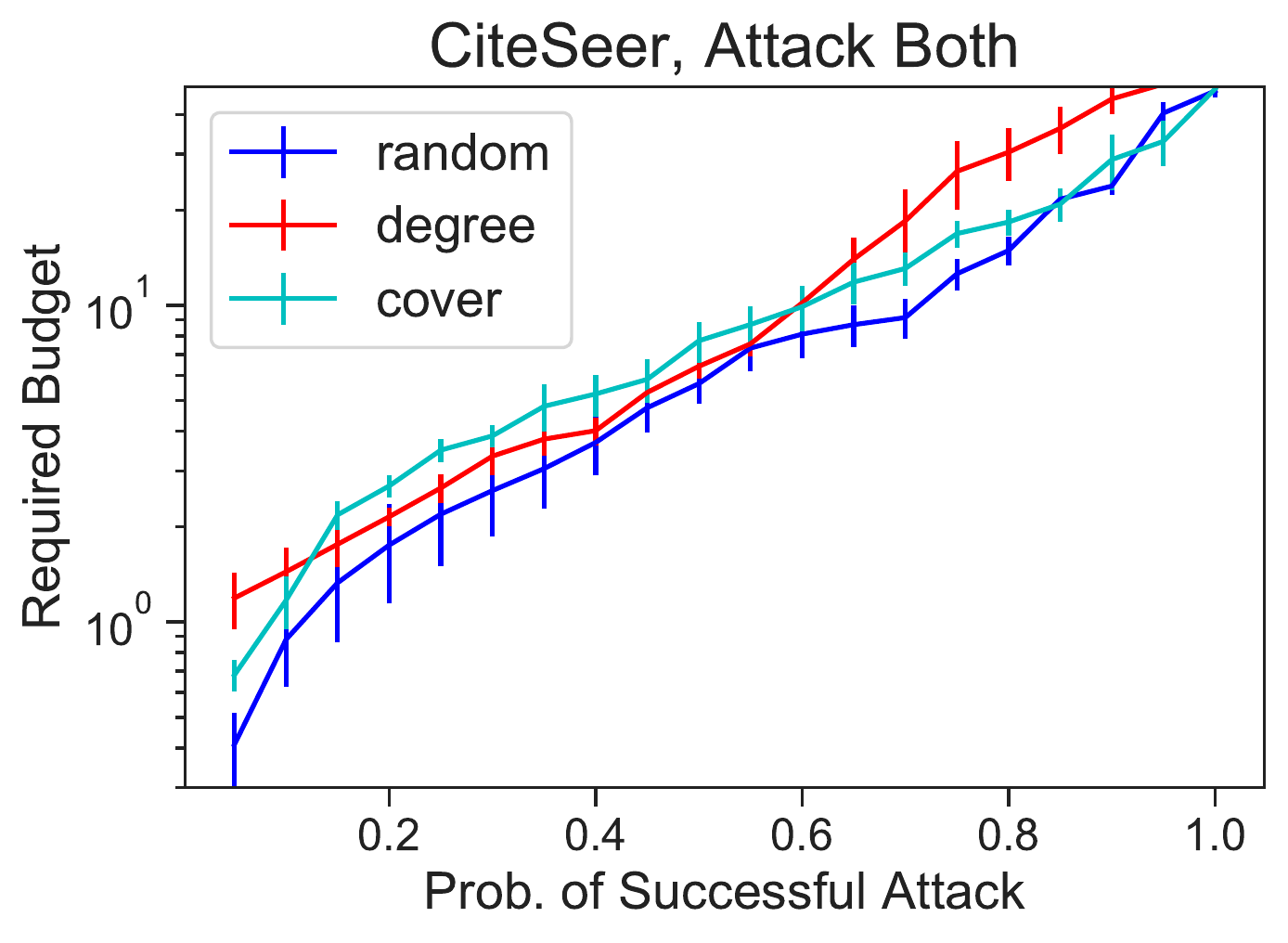}
    \caption{Budget required to achieve a given probability of attack success, varying the training data selection method. Results are shown for GCNs with one hidden layer. Curves are averages across trials and error bars show standard errors. We consider attacks on the structure of the graph (left column), the vertex attributes (center column), and attacks against both simultaneously (right column). All results use attacks against neighbors of a randomly selected target. Using \textsc{GreedyCover} consistently outperforms random selection, often by a factor of 2 and sometimes by a factor of 4. \textsc{StratDegree} also typically shows a substantial benefit.}
    \label{fig:varyTrainBudgetPlots}
\end{figure*}

In our first experiments, we apply up to 50 perturbations and attack either structure, attributes, or both. We see that, across a wide swath of attack success probabilities, the alternative selection methods provide a significant increase in robustness, often requiring a factor of 2--4 to achieve a given probability of success. Results demonstrating the increase in required budget are shown in Figure~\ref{fig:varyTrainBudgetPlots}. This figure includes influence attacks against randomly selected targets. Consider the attacks against structure alone. 
Especially in cases where the probability of attack success is low, we see both methods provide a significant increase in the adversary's required budget. In almost all cases, performance using the alternatives is appreciably better than randomly selecting training data. The increase is less pronounced in cases where attributes are attacked along with structure.

The improvement in performance is not universal across scenarios. When attributes alone are attacked, for example, the classifier is often just as robust using random training as the other methods, though the adversary's required budget is much higher. In cases where attributes and structure are both perturbed, a slightly higher budget is required for the adversary to achieve a high probability of success. Thus, the smaller performance gap between random selection and the alternative methods makes it more advantageous for the adversary to consider attributes as well as structure. Understanding this phenomenon is a goal of  ongoing work.

\begin{figure*}
    \centering
    \includegraphics[width=1.96in]{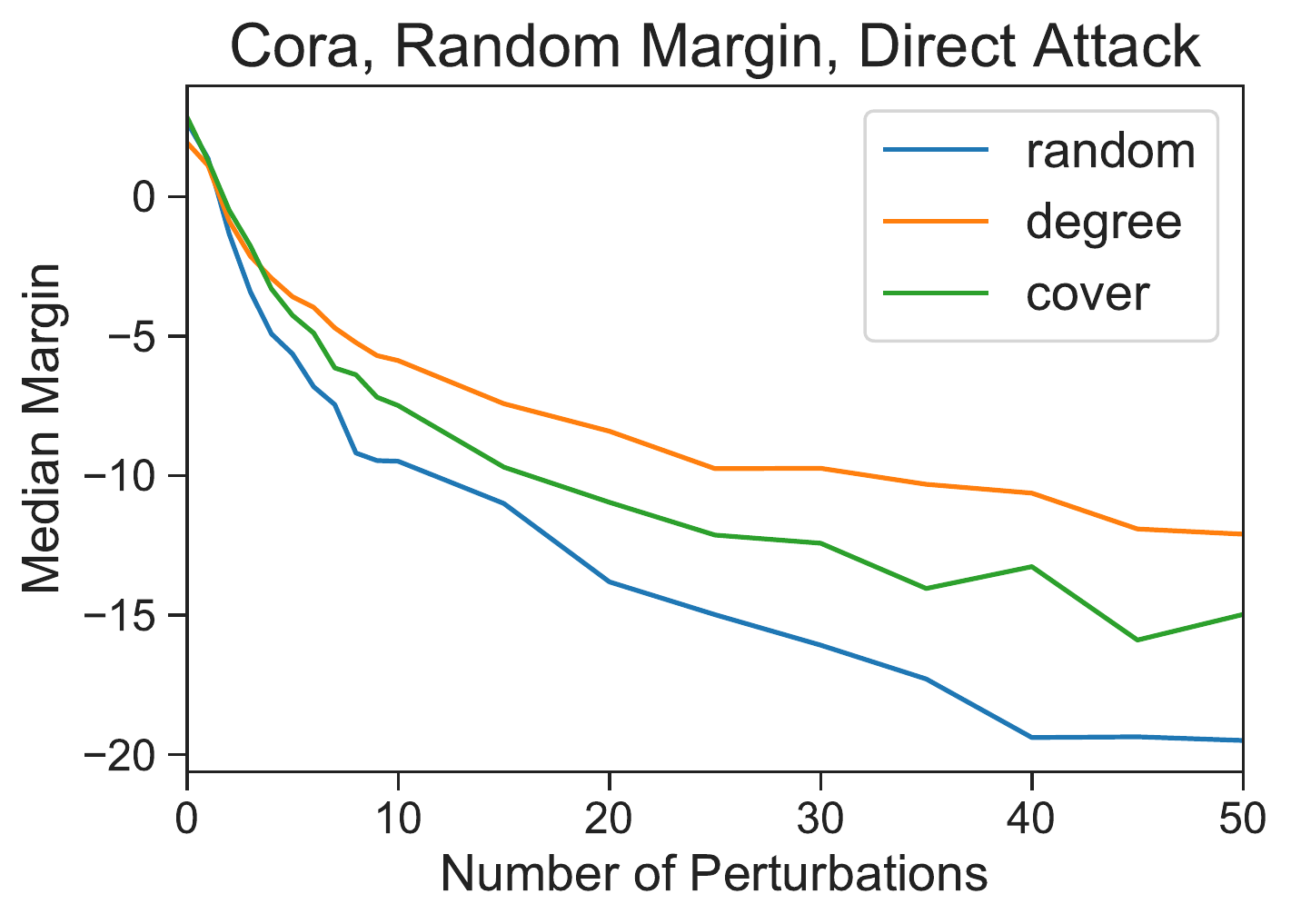}
    \includegraphics[width=1.96in]{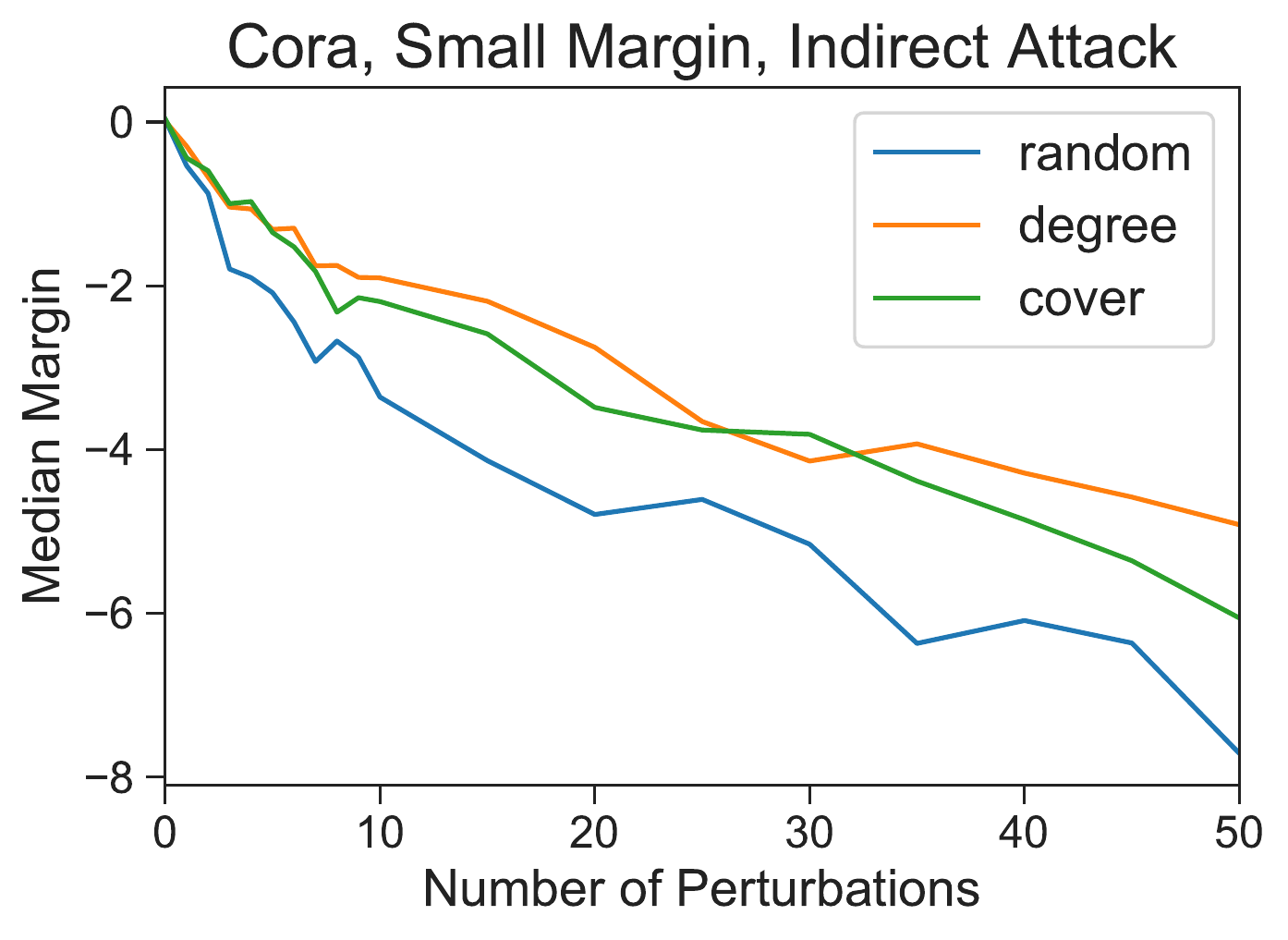}
    \includegraphics[width=1.96in]{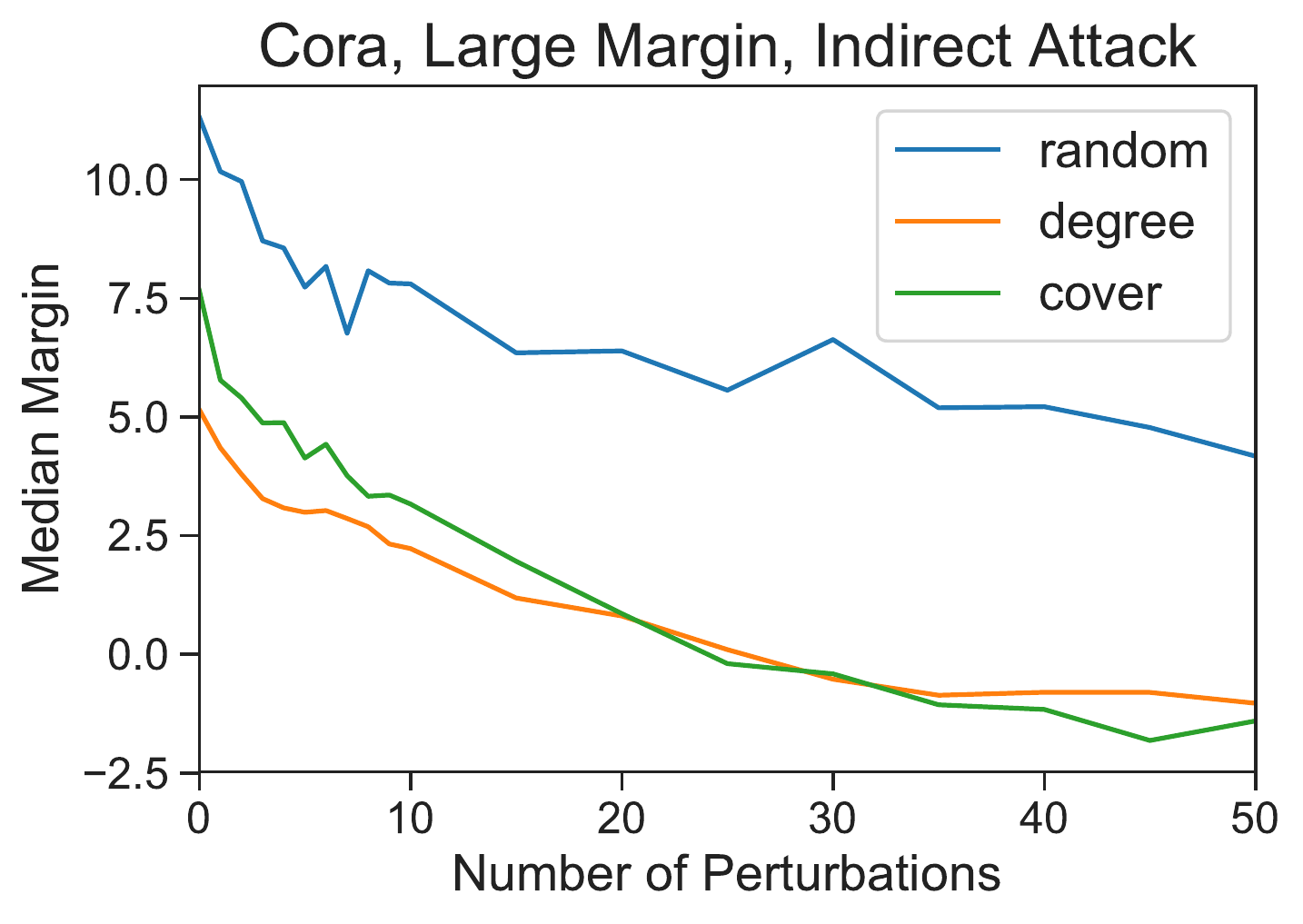}
    \caption{Median margin after perturbation, varying the training data selection method, in cases that are difficult to improve: direct attacks (left), small margin cases (center), and large margin cases (right). Attacks are against a GCN with one hidden layer and change both structure and attributes. In the direct and small margin attacks, even when misclassification occurs (0 crossing) at the same perturbation level, the margin often decreases faster with random selection.}
    \label{fig:varyTrainHardCases}
\end{figure*}

We also see low-level improvements that ultimately do not change the rate of attack success. In some cases, it is exceptionally easy or difficult for an adversary to succeed, and the varied training method does not appreciably change the number of perturbations for the attacker to succeed. We do, however, see a significant change in how much the margin is reduced, as shown for the Cora dataset in Figure~\ref{fig:varyTrainHardCases}. We see that, although the margin may cross zero at approximately the same point, the decrease is much more gradual in most cases using the alternative selection methods (direct attacks being an exception).For example, in attacks against small-margin targets, creating the same reduction in margin as with 10 perturbations under random selection requires about 20 perturbations using the alternative methods. We see that the large margin cases are \emph{much} larger using the baseline method than the alternatives, which makes those targets less vulnerable to attack, even if their margins decrease more quickly as they are perturbed. Results with CiteSeer are similar (though performance against direct attacks does not change much with \textsc{StratDegree} or \textsc{GreedyCover}) and are omitted. 

\begin{figure*}
    \centering
    \includegraphics[width=1.96in]{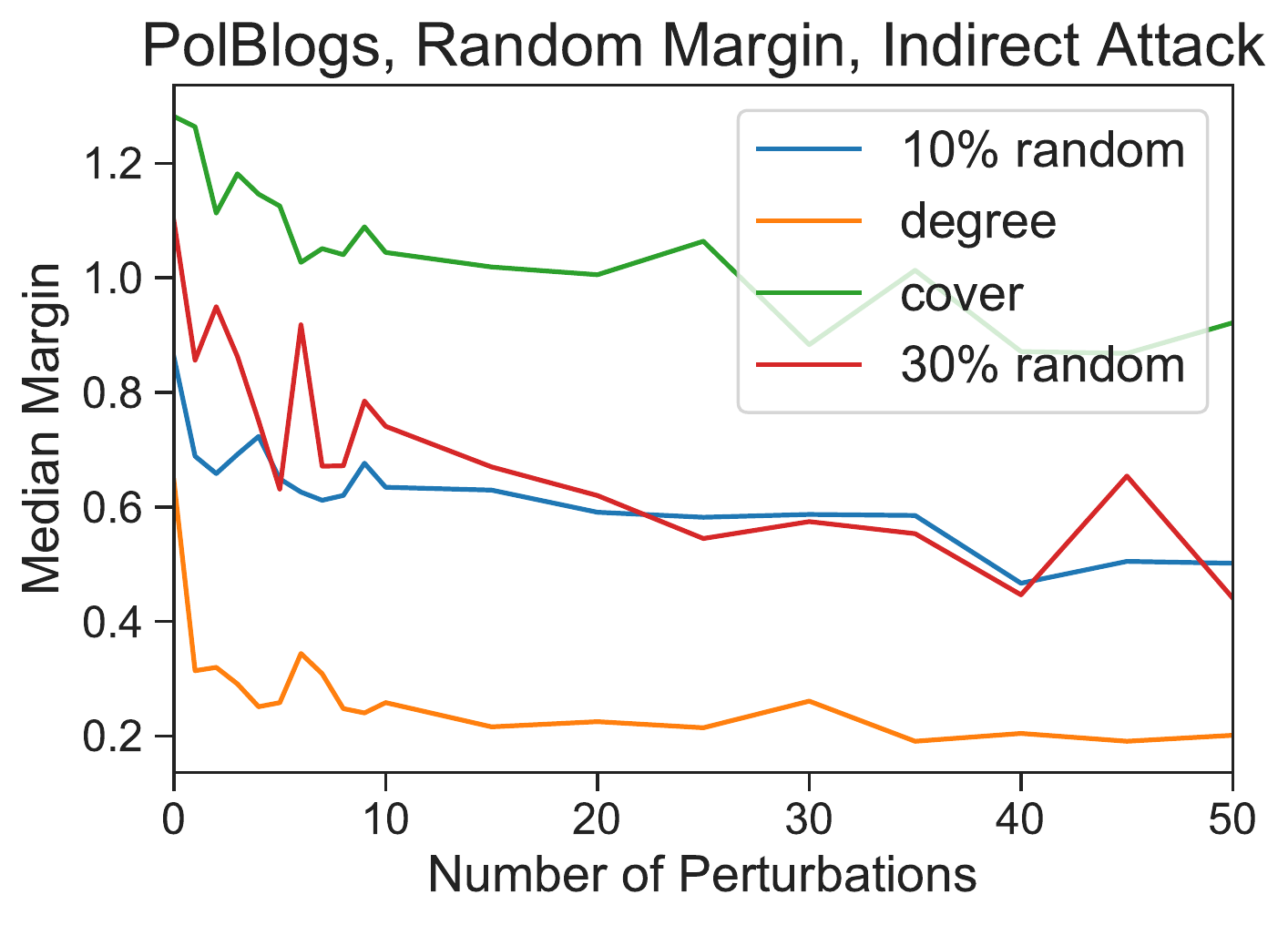}
    \includegraphics[width=1.96in]{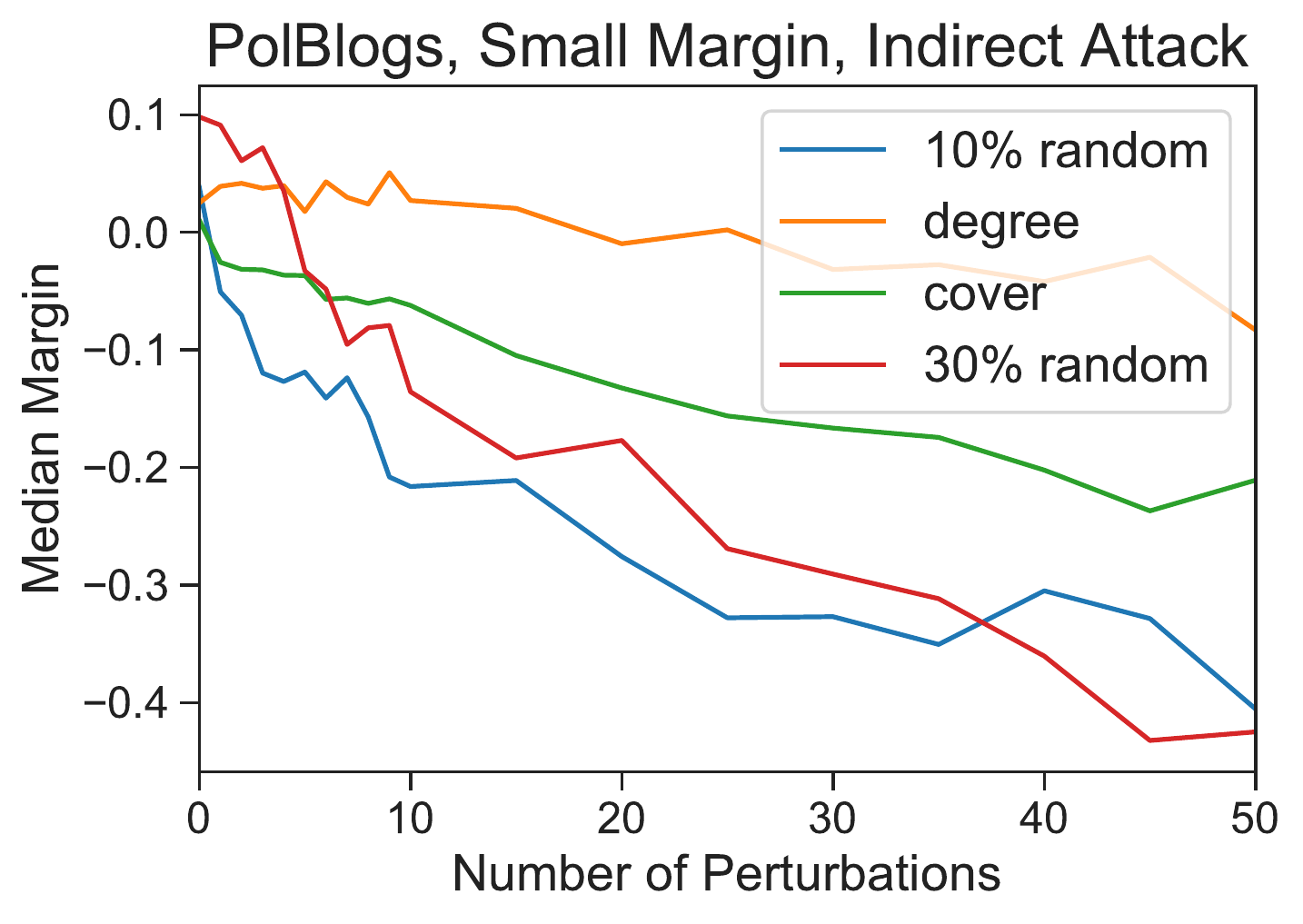}
    \includegraphics[width=1.96in]{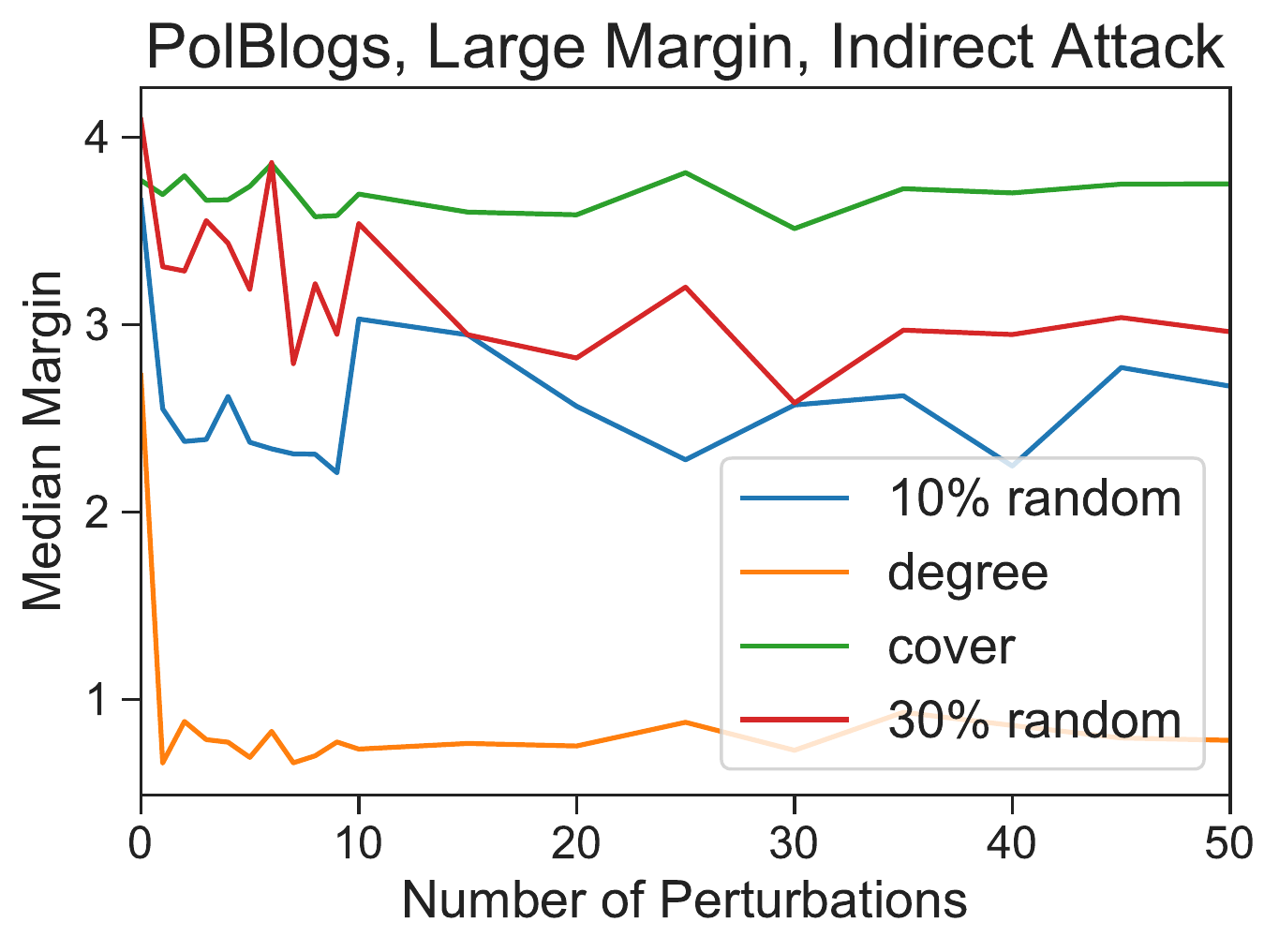}
    \caption{Median margin after perturbation, varying the training data selection method, in the PolBlogs dataset. Attacks change  structure only. In all cases, \textsc{GreedyCover} results in perturbations being less effective.}
    \label{fig:varyTrainPolBlogs}
\end{figure*}

The graph of political blogs is much more difficult to attack, and typically requires more than 50 perturbations for an attack to be successful. We show the median classification margin for up to 50 perturbations in Figure~\ref{fig:varyTrainPolBlogs}. The behavior of \textsc{StratDegree} is inconsistent (sometimes worse and sometimes better than random), but \textsc{GreedyCover} is consistently better than random, even when increasing the amount of training data to 30\%. This is true for randomly selected targets as well as those starting with very high and very low margins. The comparatively poor performance of \textsc{StratDegree} for random and large-margin targets could be due to the strong community structure in that graph, which high-degree nodes may not reveal.

\begin{figure*}
    \centering
    \includegraphics[width=1.96in]{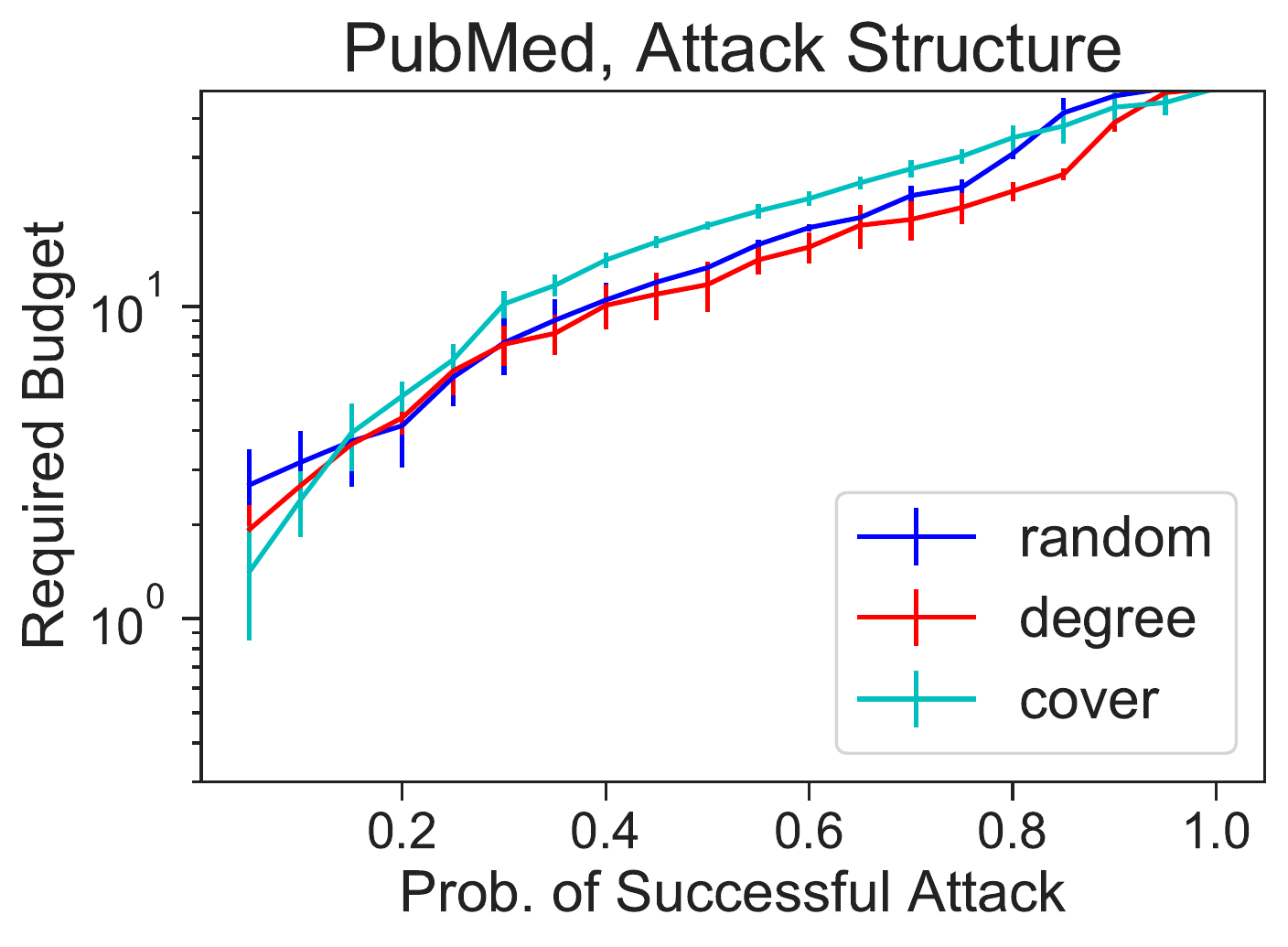}
    \includegraphics[width=1.96in]{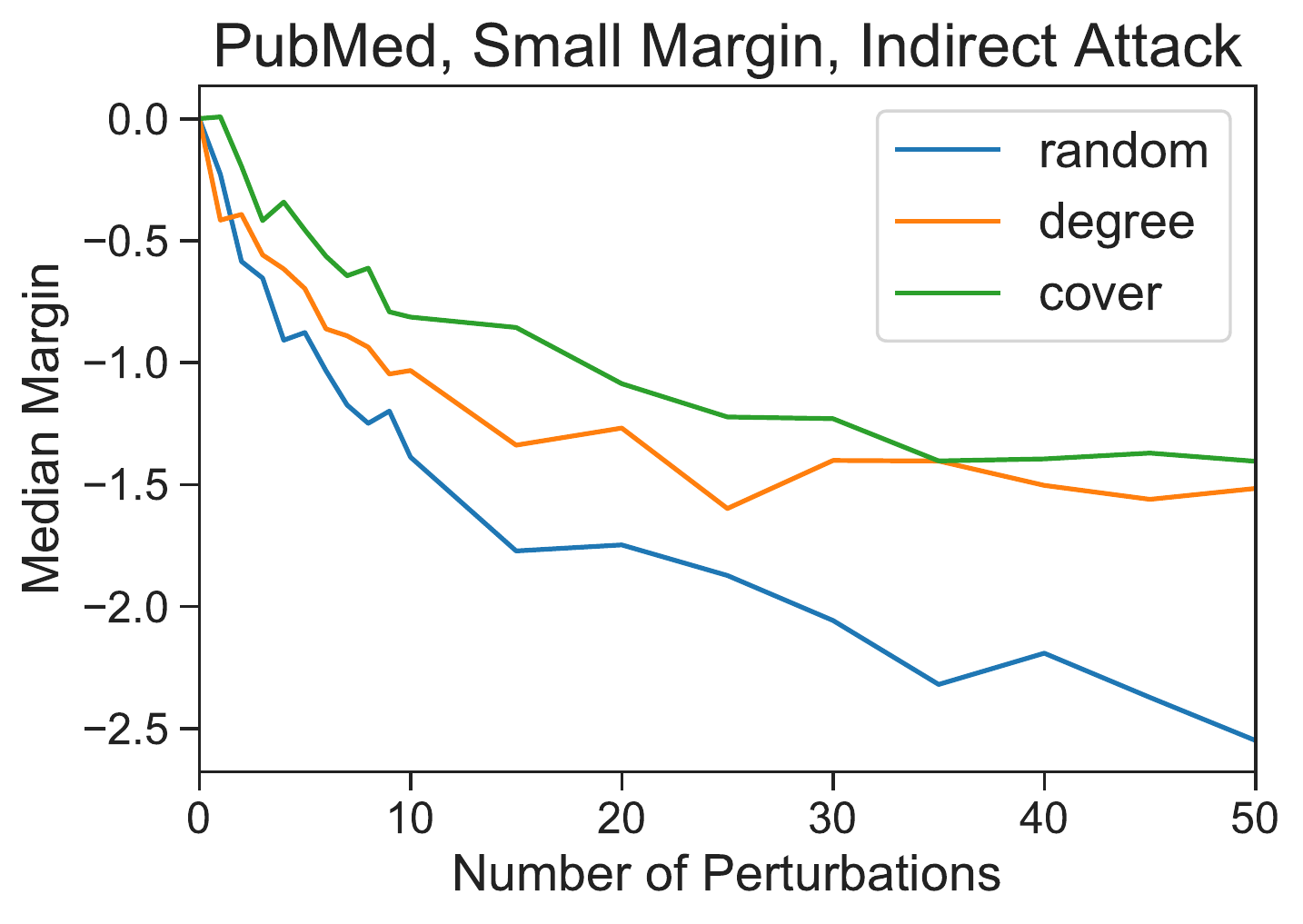}
    \includegraphics[width=1.96in]{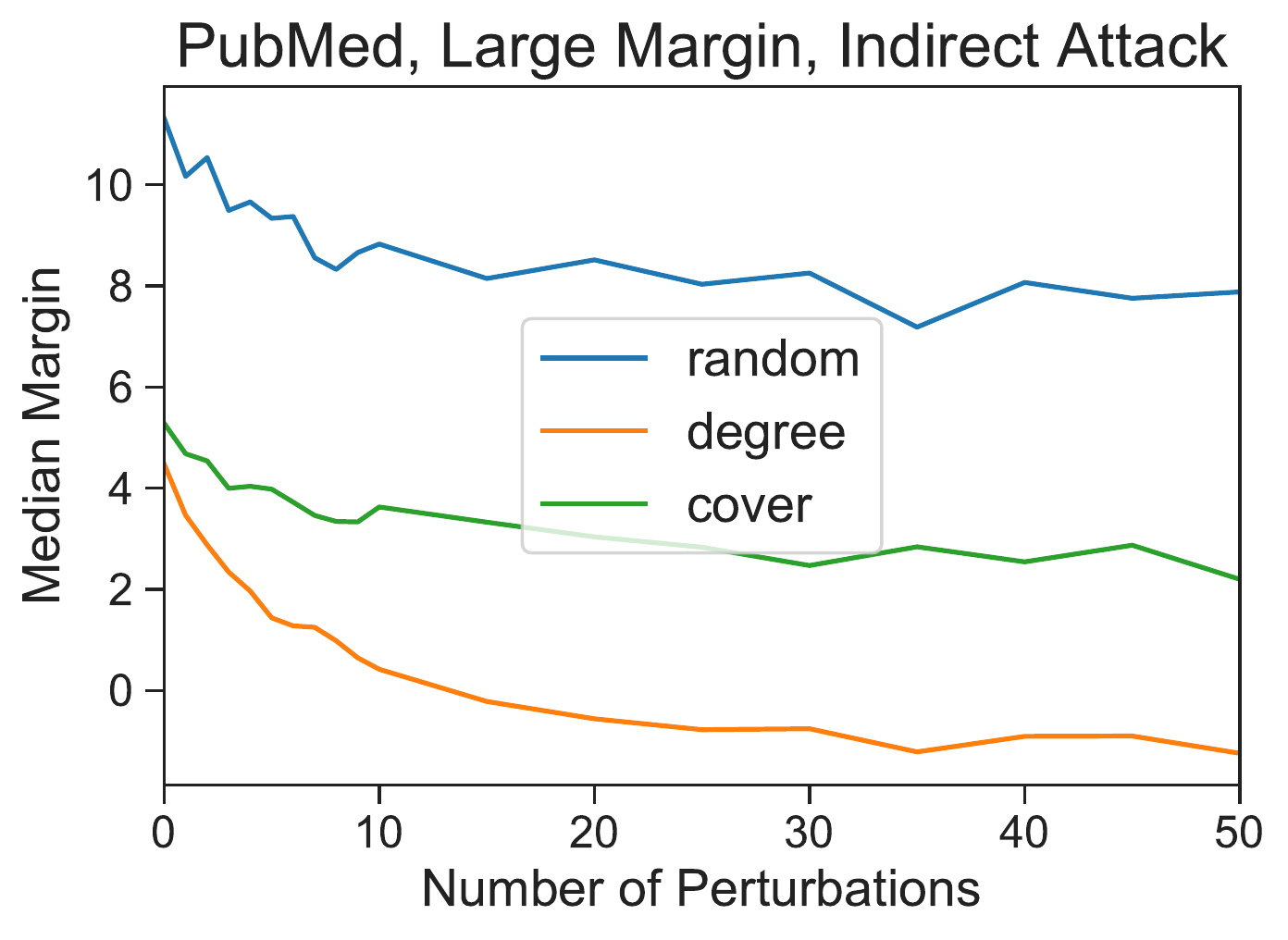}
    \caption{Results on the PubMed dataset. \textsc{GreedyCover} provides some increase in the adversary's required budget when the attack success probability is around 30\%--70\%. For cases where the margin is very  low or high, the alternative methods slow down the reduction in margin caused by the attack.}
    \label{fig:pubMed}
\end{figure*}

The PubMed dataset is much larger and we limited our analysis to indirect attacks on the graph structure. As shown in Figure~\ref{fig:pubMed}, \textsc{GreedyCover} provides a small but reliable increase in the adversary's required budget for attack success probabilities between about 0.3 and 0.7. Again,  when we consider the median classification margin, we see that training using either high degree nodes or \textsc{GreedyCover} makes the initial classification margin smaller, but slows down the rate at which the margin is reduced by  Nettack. This is shown in both small and large margin cases. In the case where the margin is large, there is a substantial decrease in the classification margin for unperturbed data, especially when using \textsc{StratDegree}. This, again, is likely due to high-degree nodes no longer being part of the unlabeled test data.

\subsection{Impact of Labeled Neighbors}
As mentioned in Section~\ref{sec:setup}, we are interested in whether the increase in robustness comes entirely from an increase in the average number of neighbors in the training set, or if there is something different about using \textsc{StratDegree} or \textsc{GreedyCover}. Thus, we compare the required adversary budget using the proposed methods to the same with more randomly selected training data. Results are shown in Table~\ref{tab:varySize}. For clarity, we use cases where only structure is attacked via influencers, where Figure~\ref{fig:varyTrainBudgetPlots} illustrates a clear empirical benefit from both methods. Using \textsc{StratDegree} and \textsc{GreedyCover} on CiteSeer, a target's average number of neighbors in the training set is $0.802$ and $0.849$, respectively, which falls in between the value for random sampling of 20\% and 25\% of the data ($0.6975$ and $0.865$, respectively).
At 20\% attack success probability, there is some substantial variation in the required budget as the random training set size increases, and this variation could potentially explain the increase when using the alternative methods. With 50\% and 80\% success, however, the performance remains fairly flat as the average number of neighbors in the training set increases. Thus, an increase in the amount of training data---to the point where we have the same number of connections to the average the test vertex---does not necessarily imply greater robustness, which is provided by the proposed methods.
\begin{table*}
    \centering
\resizebox{\textwidth}{!}{%
    \begin{tabular}{|l|c|c|c|c|c|c|c|c|c|}
        \hline
        Dataset & Thres. &\%ile& 10\% & 15\%& 20\% & 25\%& 30\% & \textsc{StratDegree}&\textsc{GreedyCover}\\
        \hline
        \hline
        \multirow{3}{*}{CiteSeer} & \multirow{3}{*}{0} & 20& 1.50 (0.17)
        & 2.87 (0.72)
        & 2.25 (0.40)
        & 2.95 (0.75)
        & 1.99 (0.29)
        & \textbf{4.02} (0.56)
        & 3.84 (0.47)\\
        \cline{3-10}
        & & 50& 3.39 (0.53)
        & 6.88 (0.81)
        & 6.65 (0.53)
        & 6.14 (1.01)
        & 6.68 (1.06)
        & \textbf{10.88} (2.00)
        & 8.69 (1.61)\\
        \cline{3-10}
        & & 80
        & 12.53 (2.04)
        & 19.75 (2.63)
        & 15.01 (1.70)
        & 17.75 (2.61)
        & 14.89 (2.61)
        & \textbf{41.04} (4.50)
        & 27.86 (4.34)\\
        \hline
        \multirow{3}{*}{Cora} & \multirow{3}{*}{0}  &20& 2.00 (0.36)
        & 2.27 (0.35)
        & 2.90 (0.70)
        & 2.08 (0.17)
        & 3.44 (0.44)
        & \textbf{3.50} (0.71)
        & 3.00 (0.53)\\
        \cline{3-10}
        & & 50 & 5.22 (0.85)
        & 5.64 (0.12)
        & 9.19 (1.36)
        & 5.48 (0.47)
        & 8.52 (0.87)
        & \textbf{10.92} (2.03)
        & 9.20 (1.17)\\
        \cline{3-10}
        & & 80& 13.84 (1.24)
        & 13.64 (1.01)
        & 18.25 (1.76)
        & 14.88 (1.15)
        & 21.45 (2.50)
        & \textbf{29.34} (6.36)
        & 17.02 (2.71)\\
        \hline
        \multirow{3}{*}{PolBlogs} & \multirow{3}{*}{$\ln{1.5}$}&20& 1.13 (0.63)
        & 1.90 (1.09)
        & 1.07 (0.46)
        & 2.69 (0.64)
        & \textbf{3.76} (2.05)
        & 0.28 (0.15)
        & 1.72 (0.86)\\
        \cline{3-10}
        & & 50& 41.89 (5.03)
        & 20.00 (10.95)
        & 22.95 (8.89)
        & 34.51 (8.82)
        & 40.16 (6.31)
        & 0.71 (0.29)
        & \textbf{50.00} (0.00)\\
        \cline{3-10}
        & & 80& \textbf{50.00} (0.00)
        & 40.32 (8.66)
        & 30.00 (10.95)
        & 40.77 (8.26)
        & \textbf{50.00} (0.00)
        & 34.87 (8.97)
        & \textbf{50.00} (0.00)\\
        \hline
    \end{tabular}%
    }
    \caption{Budget required for an adversary to achieve a given level of attack success, varying training set selection method and amount of training data. Values are the average (with standard error) number of perturbations required for the attacker to succeed the specified proportion of the time. While more randomly selected training data typically improves performance, it does not do so consistently, nor does it reliably perform better than \textsc{StratDegree} and \textsc{GreedyCover}. Note that 50 perturbations is the maximum attempted in all cases, and assigned as the required budget when no level of perturbation succeeds.}
    \label{tab:varySize}
\end{table*}

For Cora, the analogous case is not as conclusive---with random sampling in some cases coming close to the results using \textsc{StratDegree}---but it is noteworthy that the required budget does not monotonically increase with training set size. Thus, increasing the amount of randomly selected training data does not reliably improve performance. In this case, \textsc{StratDegree} ($1.135$ training neighbors per random target) and \textsc{GreedyCover} ($1.084$) yield average training neighbor rates between random sampling of 25\% ($1.029$) and 30\% ($1.237$) of the data.

PolBlogs, as mentioned in Section~\ref{subsec:varyTrain}, is much more difficult to attack, so we lower the bar for the attacker: we consider a threshold of $\ln{1.5}$ rather than 0, so rather than misclassifying the target, the probability of it belonging in the correct class is no more than 50\% higher than in any incorrect class. (Since PolBlogs has only two classes, this means the probability of being in the correct class is less than 0.6.) While neither method does better than random at low attack success rates, \textsc{GreedyCover} saturates at 50\% success (50 is the largest number of perturbations used, so anything not successfully attacked is assumed to require 50). Random sampling sometimes also saturates at a high attack success rate. For PolBlogs, \textsc{StratDegree} yields an average of $8.735$ training neighbors per random target, more than using random selection of 30\% ($8.046$), while \textsc{GreedyCover} ($6.528$) yields a rate between using 25\% ($6.708$) and 30\%.

\subsection{Impact of Defense and Adaptation}
\begin{figure*}
    \centering
    \includegraphics[width=1.96in]{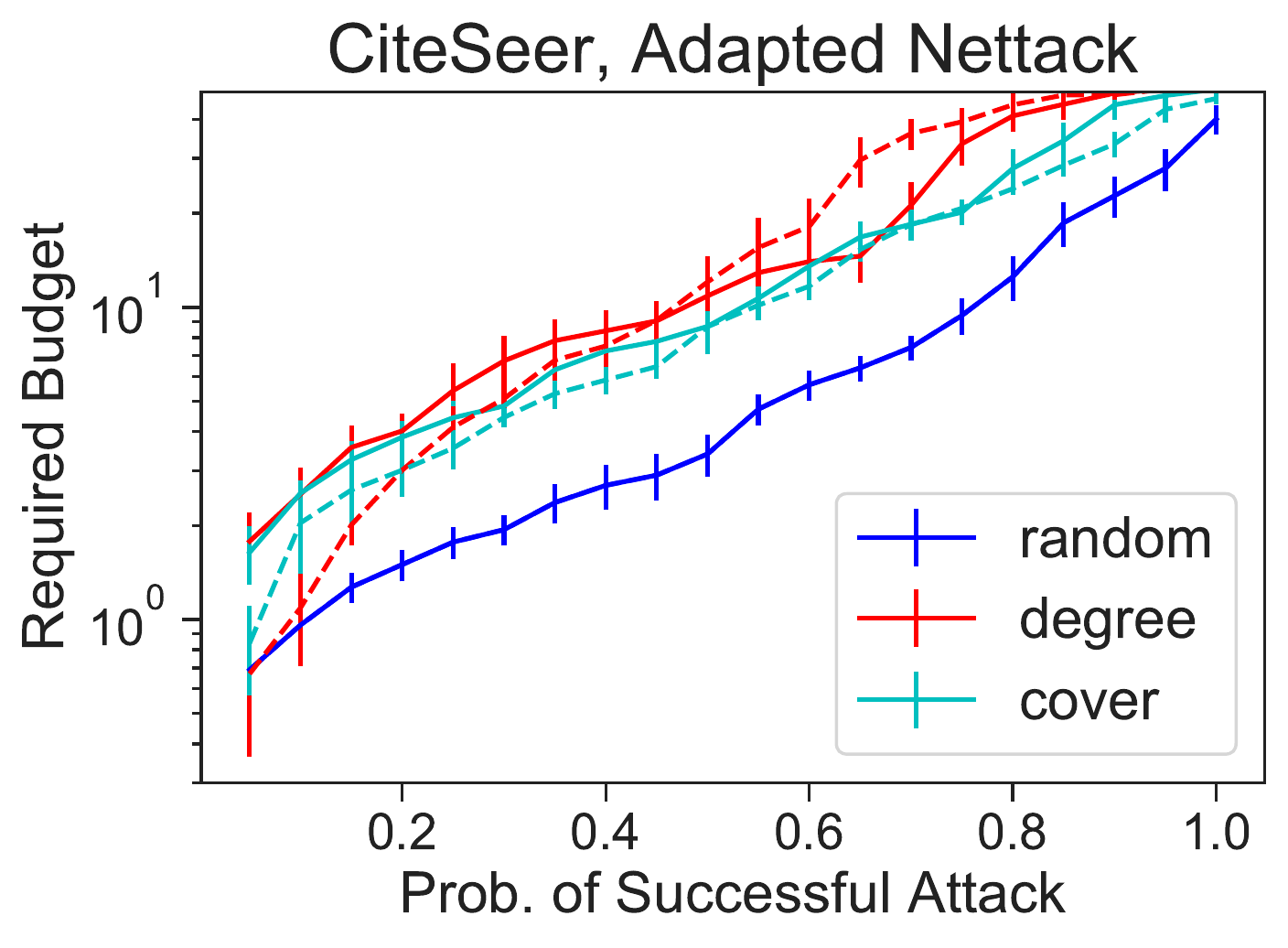}\ 
    \includegraphics[width=1.96in]{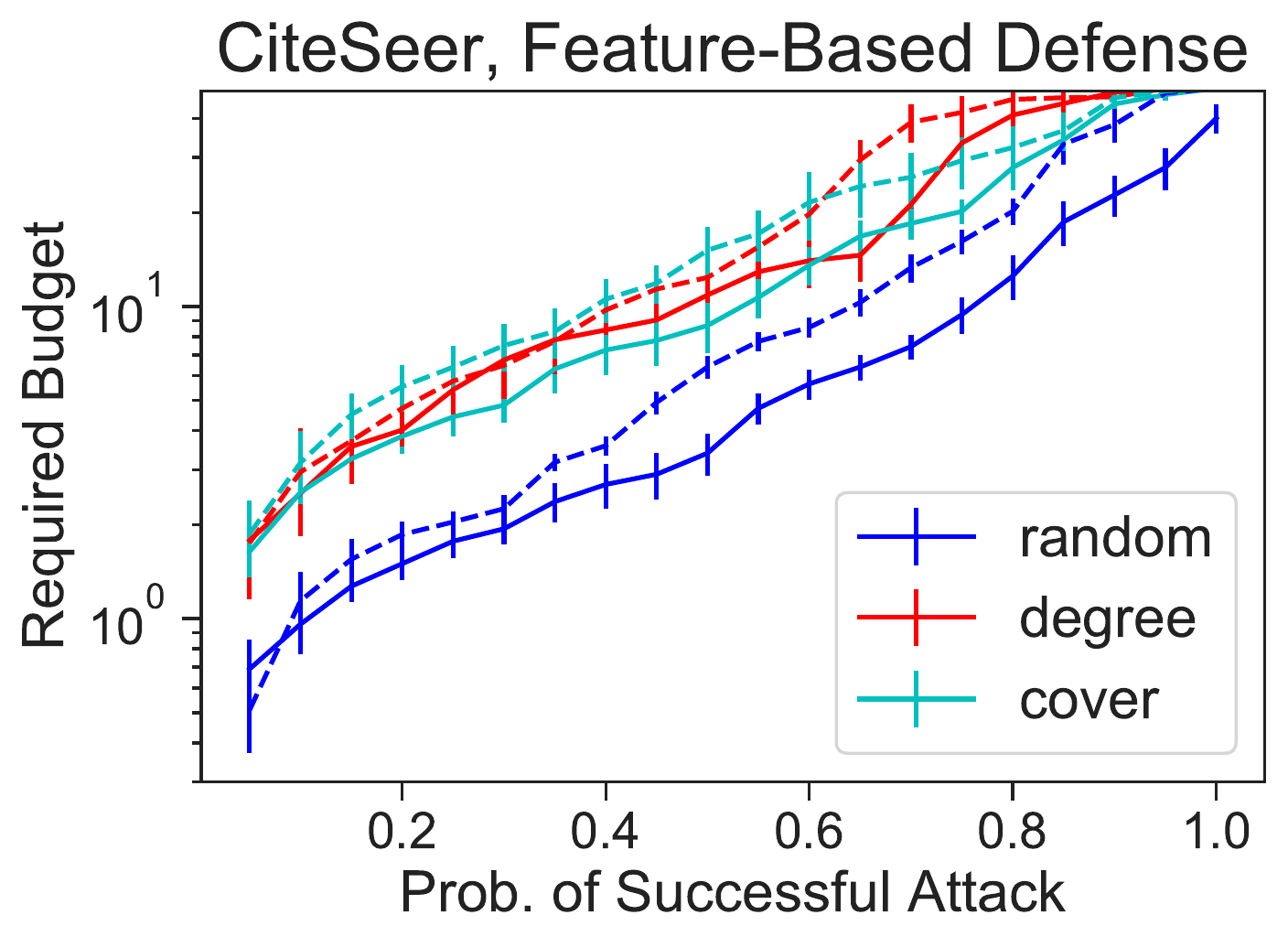}\ 
    \includegraphics[width=1.96in]{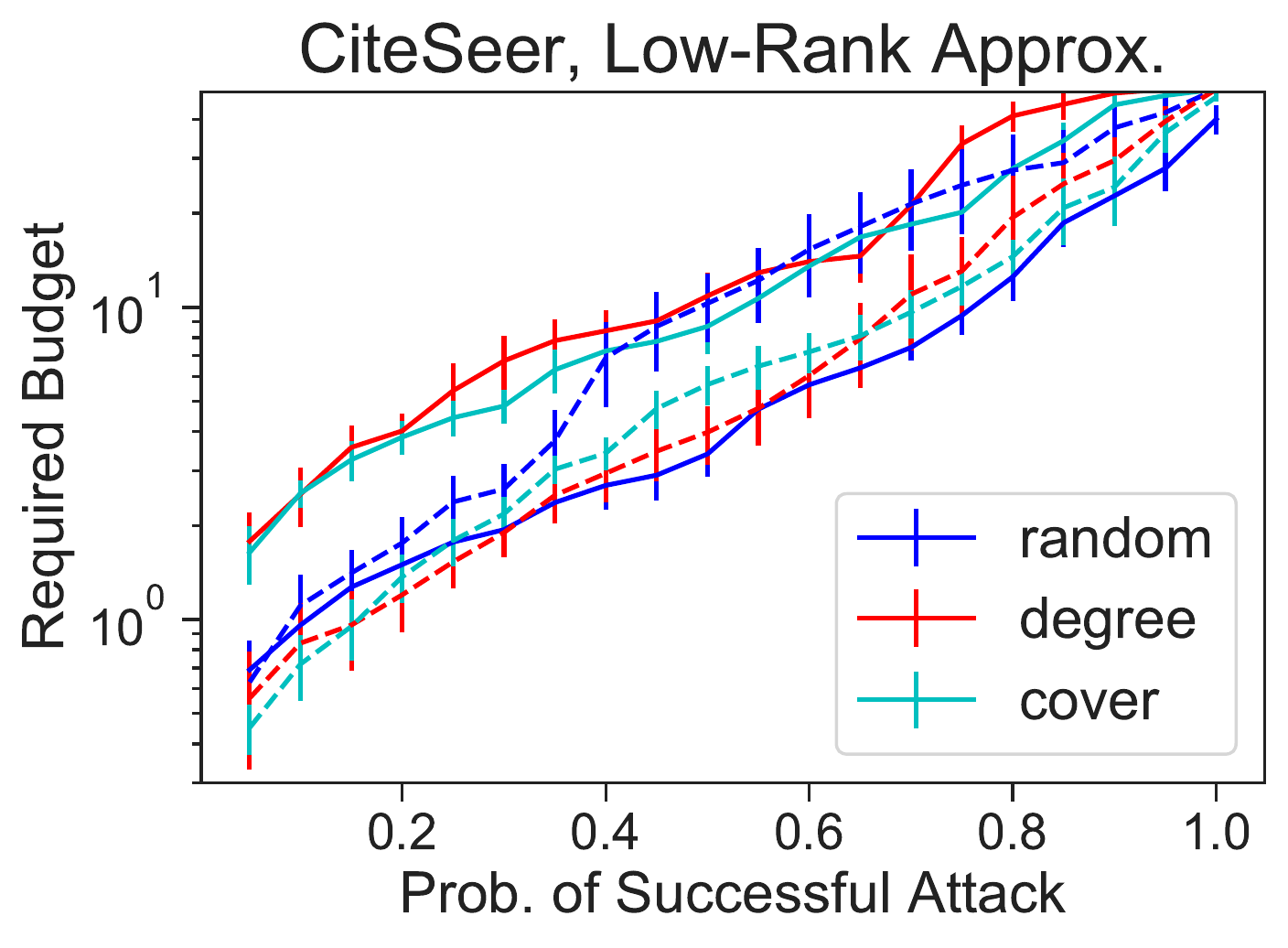}
    \caption{Results using defenses and an adapted attack on the CiteSeer dataset. Solid lines indicate the original Nettack and classifier, while dashed lines represent performance when the adaptations and defenses are employed. The adaptation of Nettack (left) does not substantially change performance and the feature-based defense (center) still benefits from the alternative methods. The low-rank defense (right) receives no benefit from the alternative methods, though at low attack success probability they are more effective than the low-rank method and are competitive (without the defense) at higher success rates.}
    \label{fig:defenses}
\end{figure*}

We applied the defenses of~\cite{Wu2019, Entezari2020} to both Cora and CiteSeer, and also used the adapted version of Nettack. As shown in Figure~\ref{fig:defenses}, when Nettack is adapted to avoid changing the training data set, there is not a significant change with the CiteSeer dataset; the adaptation negatively impacts performance at low probability of attack success but not elsewhere. At no point does it make the alternative methods less effective than random selection, and at most  points they  maintain the  performance achieved with no adaptation of the attack. For the feature-based defense of~\cite{Wu2019}, we see an increase in the adversary's budget using random selection, as expected, but we also see a small increase in the required budget using the alternative methods. This suggests that, even when the adversary is cautious about altering the training data, the \textsc{StratDegree} and \textsc{GreedyCover} are helpful. 

The low-rank approximation, on the other hand, does not benefit at all from the alternative methods. They are competitive with the low-rank defense and in fact provide better robustness at low probability of attack success. When using the low-rank approximation, however, both \textsc{StratDegree} and \textsc{GreedyCover} usually yield worse performance than random selection (in  fact, performance is similar to when random selection is used for the full graph). This is an intriguing phenomenon. It could be due to the fact that high-degree nodes often dominate low-rank approximations, so the alternative methods are not providing much new information, or simply due to the very locally focused nature of both \textsc{StratDegree} or \textsc{GreedyCover}, the effects of which could be filtered out by a more global method. This result suggests that these methods should not be used in conjunction with one another.

Results on the Cora dataset are similar and omitted for space.
\subsection{Possible Accuracy/Robustness Tradeoff}
\begin{figure}[t!]
    \centering
    \includegraphics[width=2.25in]{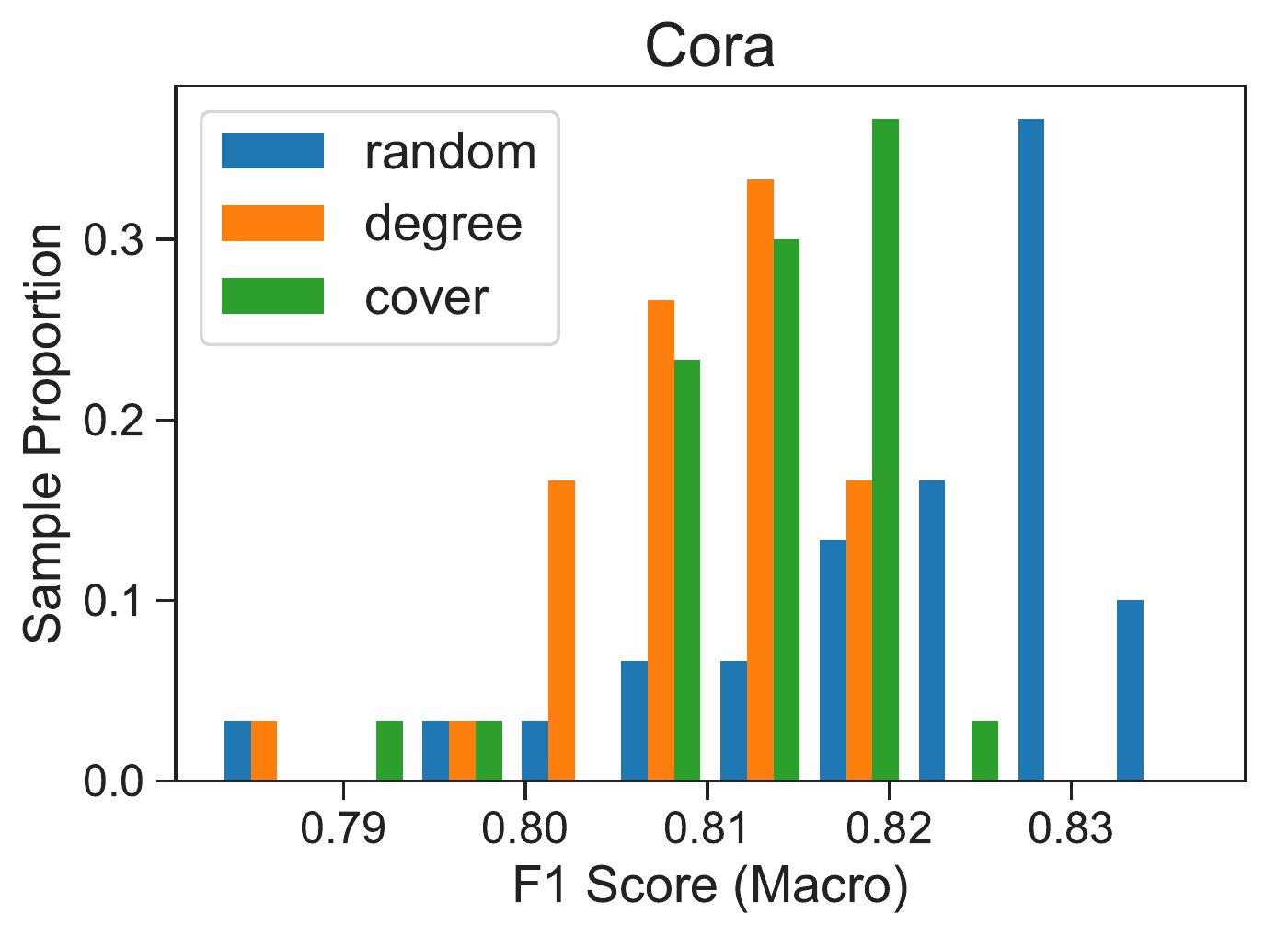}
    \includegraphics[width=2.25in]{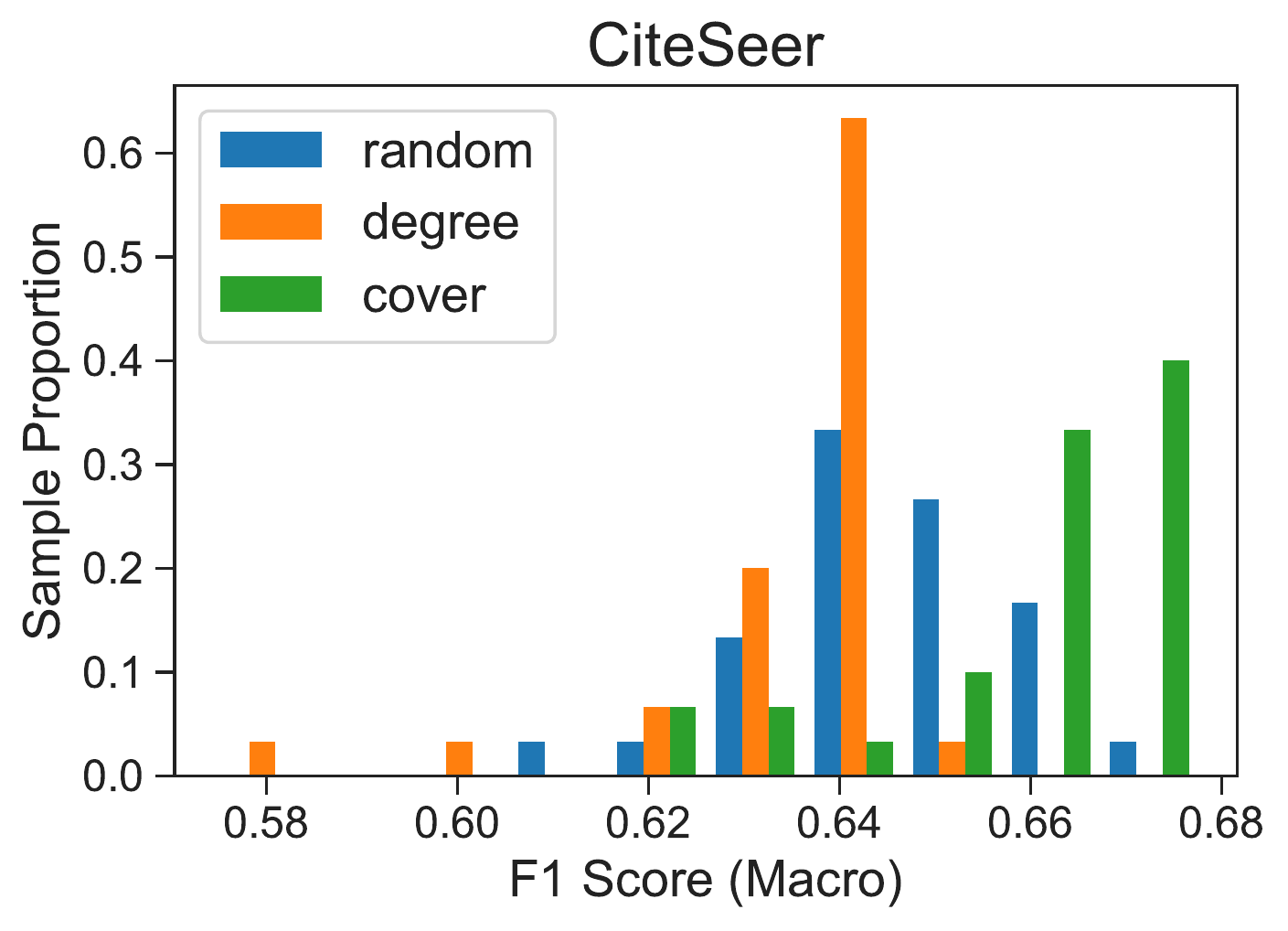}
    \caption{Distribution of F1 scores (macro aggregated) using all 3 training set selection methods on both datasets before perturbation. Performance is for a GCN with one hidden layer. The alternative methods, degree-based and greedy-cover, sometimes reduce performance, suggesting the existence of a tradeoff between robustness and classification performance.}
    \label{fig:varyTrainEval}
\end{figure}

One additional consideration is whether either of these methods degrades overall system performance, thus yielding a tradeoff between overall performance and robustness. The experiments we have run so far suggest that there is some variation, as illustrated in Figure~\ref{fig:varyTrainEval}. While it seems that \textsc{StratDegree} does slightly degrade classification performance, it is less clear that the \textsc{GreedyCover} does; it slightly improves performance for CiteSeer. Performance for PolBlogs is similar to Cora, reducing the F1 score from approximately 0.952 with random training to 0.939 for \textsc{StratDegree} and 0.911 for \textsc{GreedyCover}. PubMed sees a mild decrease in performance, from 0.855 with random to 0.843 with \textsc{StratDegree} and 0.852 for \textsc{GreedyCover}.

We considered the possibility that this disparity could be due to importance of structure versus attributes. To explore this, we decoupled structure from attributes and considered an alternative classifier in which the graph is embedded into 128-dimensional Euclidean space using node2vec~\cite{Grover2016}, the resulting vector is appended to the node attributes, and the augmented feature vector is classified using a support vector machine with a radial basis function kernel. Results are shown in Table~\ref{tab:decoupled}. Note that classification performance on Cora is much better when ignoring attributes than when ignoring structure, while classification is easier for CiteSeer when only attributes are used than using structure alone. It could be that the greater dependence on structure makes classification performance more sensitive to selecting training data based on structure. This is an important area for future investigation.
\begin{table*}
    \centering
    \begin{tabular}{|l|c|c|c|c|c|}
        \hline
         \multirow{ 2}{*}{Dataset}& \multicolumn{2}{c|}{GCN} & \multicolumn{3}{c|}{SVM}\\
         \cline{2-6}
         &w/o attributes &with attributes & node2vec alone & attributes alone &node2vec + attributes\\
         \hline
         Cora & 0.8560 (0.028)& 0.8785 (0.026)& 0.8623 (0.024)& 0.7471 (0.022)& 0.8675 (0.025)\\
         \hline
         CiteSeer & 0.7063 (0.021)& 0.7585 (0.021)& 0.7071 (0.035)& 0.7237 (0.022)&0.7503 (0.033)\\
         \hline
    \end{tabular}
    \caption{Vertex classification performance with different classifiers and different features. Mean accuracy (with standard deviation) across 10 trials is shown. Cora is much more easily classified with structure than attributes, while CiteSeer is more easily classified by attributes.}
    \label{tab:decoupled}
\end{table*}

\subsection{Simulation Results}
\begin{figure}
    \centering
    \includegraphics[width=2.25in]{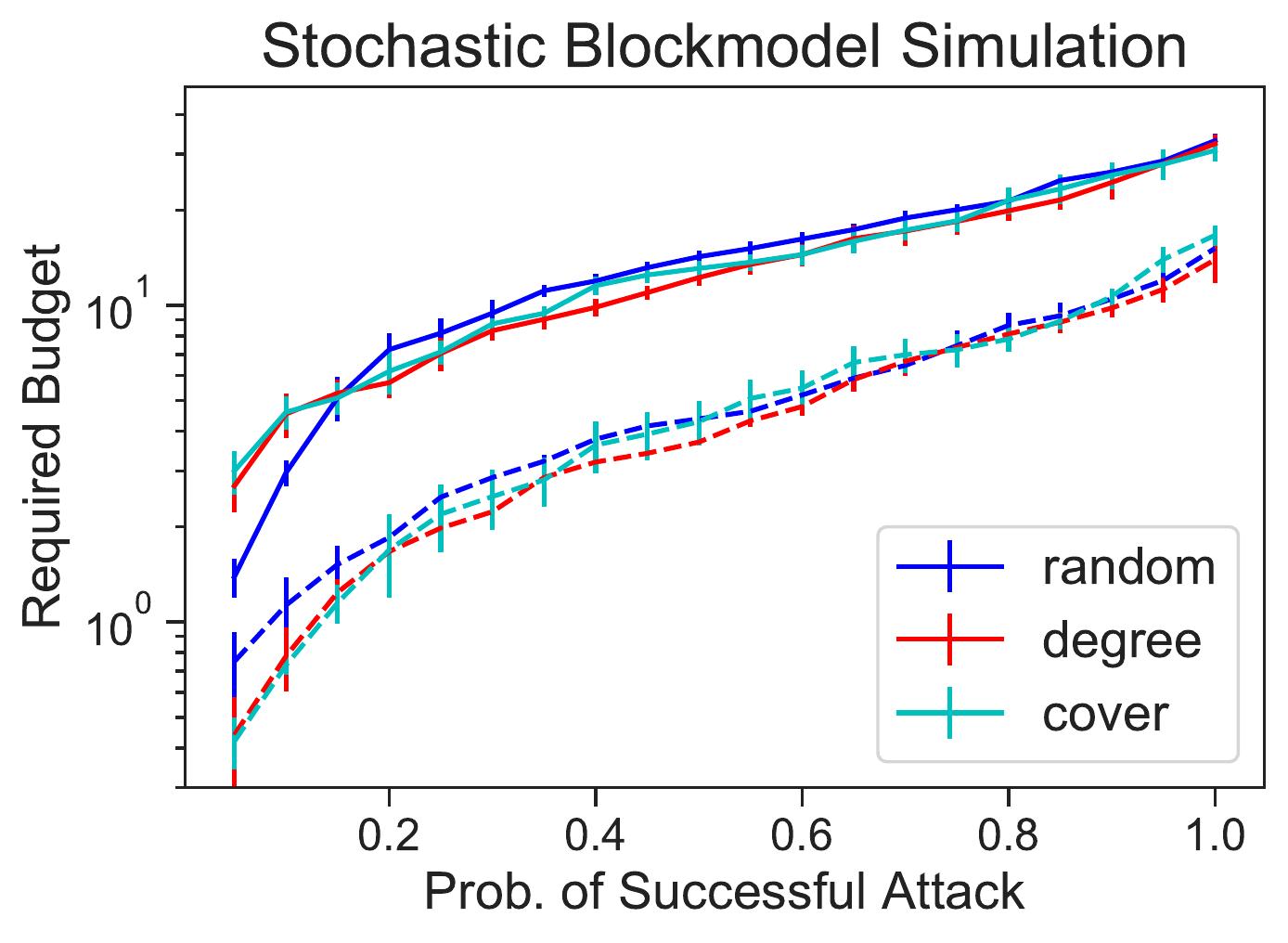}
    \caption{Results of a stochastic blockmodel simulation. The solid line is a case with greater within-community connectivity than the dashed line. The proposed selection techniques, \textsc{StratDegree} and \textsc{GreedyCover}, are only effective when the probability of attack success is low. That is, the more  heterogeneous the degree distribution of the network, the more benefit our proposed techniques produce. Note that most real-world networks have heterogeneous degree distributions.}
    \label{fig:simulation}
\end{figure}

In an effort to understand what makes the proposed methods effective, we test them in a controlled setting on a simulated graph. We use a stochastic blockmodel with 5 classes, 400 nodes per class, where classes are aligned with blocks. Simulation details are provided in the supplement. As with the real data, we ran 5 trials of the same experiments, in this case generating a  new graph for each trial (rather than only changing the train/test/validation split). As shown in Figure~\ref{fig:simulation}, the proposed techniques are only effective at very low probability of attack success. This suggests that both degree-based selection and \textsc{GreedyCover} rely on more heterogeneous degree distributions to provide a benefit to the user. Investigating this in more detail will be an aspect of future work.

\section{Related Work}
\label{sec:related}
Adversarial examples in deep neural networks have received considerable attention since they were documented a few years ago~\cite{Szegedy2014}. Since that time, numerous attack methods have been proposed, largely focused on the image classification domain (though there has been interest in natural language processing as well, e.g.,~\cite{Jia2017}). In addition to documenting adversarial examples, Szegedy et al. demonstrated that such examples can be generated using the limited-memory BFGS (L-BFGS) algorithm, which identifies an adversarial example in an incorrect class with minimal $L_2$ norm to the true data. Later, Goodfellow et al. proposed the fast gradient sign  method (FGSM), where the attacker starts with a clean image and takes small, equal-sized steps in each dimension (i.e., alters each pixel by the same amount) in the direction maximizing the loss~\cite{Goodfellow2015}. Another proposed attack---the Jacobian-based Saliency Map Attack (JSMA)---iteratively modifies the pixel with the largest impact on the loss~\cite{Papernot2016a}. DeepFool, like L-BFGS, minimizes the $L_2$ distance from the true instance while crossing a boundary into an incorrect class, but does so quickly by approximating the classifier as linear, stepping to maximize the loss, then correcting for the true classification surface~\cite{Moosavi-Dezfooli2016}. Like Nettack, these methods all try to maintain closeness to the original data ($L_2$ norm for L-BFGS and DeepFool, $L_0$ norm for JSMA, and $L_\infty$ norm for FGSM).

The Nettack paper only compares to one of these methods: FGSM. In~\cite{Wu2019}, the authors modify JSMA to use integrated gradients and show it to be effective against vertex classification. 
In addition, new attacks against vertex classification have been introduced, including a method that uses reinforcement learning to identify modifications to graph structure for an evasion attack~\cite{Dai2018} and using meta learning for a global (non targeted) attack~\cite{Zugner2019a}. Exploring the effects of \textsc{StratDegree} and \textsc{GreedyCover} on these methods will be interesting future work.

Defenses to attacks such as FGSM and JSMA have been proposed, although several prove to be insufficient against stronger attacks. Defensive distillation is one such defense, in which a classifier is trained with high ``temperature'' in the softmax, which is reduced for classification~\cite{Papernot2016}. While this was effective against the methods from~\cite{Szegedy2014,Goodfellow2015, Papernot2016a,Moosavi-Dezfooli2016}, it was shown in~\cite{Carlini2017} that modifying the attack by changing the constraint function (which ensures the adversarial example is in a given class) renders this defense ineffective. More defenses have been proposed, such as pixel deflection~\cite{Prakash2018} and randomization techniques~\cite{Xie2018}, but many such methods are still found to be vulnerable to attacks~\cite{Athalye2018,Athalye2018a}. Future work will also consider defenses that appear promising (e.g.,~\cite{Wong2018, Croce2019}) if they can be applied in the graph domain. More recent work has focused on robustness of GCNs, including work on robustness to attacks on attributes~\cite{Zugner2019b} and more robust variants of the GCN~\cite{Zhu2019}. In particular, evaluating the impact of \textsc{StratDegree} and \textsc{GreedyCover} on certifiable robustness as outlined in~\cite{Bojchevski2019b} will be important future work. Finally, attacks against classifiers other than GCNs~\cite{Yu2018} and objectives other than vertex classification---such as node embedding~\cite{Bojchevski2019a} and community detection~\cite{Kegelmeyer2018}---would be interesting to consider, to determine whether any strategy works against many adversary tactics. 
\section{Conclusions}
\label{sec:conclusion}
This paper explores the use of complex network characteristics to make vertex classification more robust to the adversarial poisoning technique Nettack. We consider an alternative approach to enhance robustness to Nettack: 
varying the training set selection method based on connections from training to test data. By using these methods, we see an improvement in robustness, both in terms of increasing the budget required for a successful attack (often by a factor of 2 or more) and, in cases where this is not possible, making the classifier much less confident in its incorrect predictions. The level of robustness achieved from the alternative methods is often not obtainable by simply increasing the amount of training data via random selection. These developments are highly encouraging as we explore possibilities to enhance vertex classification robustness.

The work documented here points to several open problems and avenues of potential investigation. 
First and foremost, developing methods that consistently achieve the highest performance of the GCN on the overall dataset with the higher robustness is the ultimate goal, and developing a hybrid method of selecting the training data that uses a combination of randomly selected data and nodes that cover the test set could prove useful if it enables both large classification margins and robustness to perturbations. 
Adapting Nettack beyond the first-order adaptation outlined here is another important problem. A full suite of experiments with several attacks, as recommended in~\cite{Carlini2019}, would be highly valuable.
Another interesting question is whether there are certain properties of a graph that make the robustness we pursue possible. Quantifying a tradeoff between robustness and performance---in the spirit of~\cite{Tsipras2019}, focused on graph data---would help elucidate some of the phenomena documented here. 
We should also consider additional network characteristics such as the triangle distribution. For example, we observed that direct attacks from Nettack increase triangle count~\cite{Miller2019} and it will be interesting to see under what conditions this becomes a reliable attack detector. 
Robustness to adversarial activity has driven fascinating research in the image classification domain.  We look forward to new discoveries as the same is done for vertex classification.


\bibliographystyle{ACM-Reference-Format}
\bibliography{bibfile}
\clearpage\section*{Appendix on Reproducibility}

\paragraph{Stochastic block model code and parameters.}Figure~\ref{fig:sbmruns} outlines the stochastic block model code and parameters that we used. For the case with stronger connectivity within communities, within-class connection probability is set to $p_\mathrm{in}=0.0140$ and between-class connections occur with probability $p_\mathrm{out}=0.0015$ (\verb|inprob| in Figure~\ref{fig:sbmruns} is set to 0.0125). For the less internally connected case, \verb|inprob| is set to 0.00625. The nodes' 50 binary features were independent and for each class a distinct set of 10 features had probability 0.35 of being 1 while all other features had a probability of 0.1. The function \verb|loadSBM| returns the adjacency matrix, attribute matrix, and labels in the same format as the \verb|load_npz| function in the original Nettack code (available at \url{https://github.com/danielzuegner/nettack}).
\begin{figure}[hbt!]
\footnotesize
\begin{verbatim}
import numpy as np
import networkx as nx
import scipy.sparse as sp

#inprob: additional internal connection probability (difference
#with between-community probability)
def loadSBM(inprob): 
    #create graph
    N = [400, 400, 400, 400, 400]
    x = inprob
    y = .004-.2*x
    P = x*np.eye(5) + y*np.ones((5, 5))
    G = nx.stochastic_block_model(N, P)
    A = nx.adjacency_matrix(G).astype('d')

    #create attributes
    p1 = .35
    p2 = .1
    Pmat = 
       np.kron((p1-p2)*np.eye(5)+p2*np.ones(5),
                np.ones((400, 10)))
    X = np.random.rand(2000, 50) < Pmat
    X = sp.csr_matrix(X)

    z = 
       np.kron(np.arange(5), np.ones((400), 
               dtype=np.int32))
    # return adjacency matrix, attribute matrix, and classes
    return A, X, z
\end{verbatim}
\caption{Stochastic blockmodel code and parameters}
\label{fig:sbmruns}
\end{figure}

\paragraph{Modified GCN code.} Figure~\ref{fig:GCN} contains the modified GCN code that we used to implement the low-rank approximation of the adjacency matrix. This code replaces code in the Nettack implementation that starts on the line\\
\verb|            self.An = tf.SparseTensor( ...|\\
and goes to the assignment of \verb|self.h1|. In addition, the change made to the last line in Figure~\ref{fig:GCN} (replacing the $AXW_1$ term) is analogously made to the line computing \verb|self.h2|.
\begin{figure}[hbt!]
\footnotesize
\begin{verbatim}
Un, Sn, VTn = la.svds(An, k=rank)
SVTn = np.diag(Sn) @ VTn
X_U, X_S, X_VT = la.svds(X_obs, k=rank)
X_SVT = np.diag(X_S) @ X_VT
self.Un = tf.convert_to_tensor(Un, dtype=tf.float32)
self.SVTn = tf.convert_to_tensor(SVTn, dtype=tf.float32)

self.X_sparse = tf.SparseTensor(np.array(X_obs.nonzero()).T,\
                                X_obs[X_obs.nonzero()].A1,\
                                X_obs.shape)
self.X_dropout = sparse_dropout(self.X_sparse, 1 - self.dropout,
                                  (int(self.X_sparse.values.get_shape()[0]),))
dropoutArray = 
   tf.sparse.to_dense(self.X_dropout).eval(session=self.session)
X_drop_U, tempS, tempVT = la.svds(dropoutArray, k=rank)
del dropoutArray
X_drop_SVT = np.diag(tempS) @ tempVT
self.X_drop_U = tf.convert_to_tensor(X_drop_U, dtype=tf.float32)
self.X_drop_SVT = tf.convert_to_tensor(X_drop_SVT, dtype=tf.float32)

self.X_U = tf.convert_to_tensor(X_U, dtype=tf.float32)
self.X_SVT = tf.convert_to_tensor(X_SVT, dtype=tf.float32)
# only use drop-out during training
self.X_comp1 = tf.cond(self.training,
                      lambda: self.X_drop_U,
                      lambda: self.X_U) if self.dropout > 0. else self.X_U
self.X_comp2 = tf.cond(self.training,
                      lambda: self.X_drop_SVT,
                      lambda: self.X_SVT) if self.dropout > 0. else self.X_SVT

self.W1 = slim.variable('W1', 
                        [self.D, sizes[0]], tf.float32, initializer=w_init())
self.b1 = slim.variable('b1', 
                        dtype=tf.float32, initializer=tf.zeros(sizes[0]))

self.h1 = dot(self.Un, dot(self.SVTn, dot(self.X_comp1,\
                                          dot(self.X_comp2, self.W1))))
\end{verbatim}
\caption{Modified GCN code that uses a low-rank approximation rather than the sparse adjacency matrix. Note that the \texttt{la} package is \texttt{scipy.sparse.linalg}; all other packages are as in the original code.}
\label{fig:GCN}
\end{figure}

\paragraph{Similarity-based defense code.} In Figure~\ref{fig:similarityCode}, we show the implementation of the similarity-based defense. When considering this defense, any time \verb|nettack.GCN.GCN()| is called, the function \verb|removeDissimilar| is first called on the adjacency matrix and attributes to remove any edges connecting nodes with no feature overlap.
\begin{figure}[hbt!]
\footnotesize
\begin{verbatim}
import scipy.sparse as sp
def removeDissimilar(A, X):
    r, c = sp.triu(A).nonzero()
    Xd = X.todense()
    for ii in range(len(r)):
        if np.dot(Xd[r[ii]], np.transpose(Xd[c[ii]])) == 0:
            A[r[ii], c[ii]] = 0
            A[c[ii], r[ii]] = 0
\end{verbatim}
\caption{Python function implementing a similarity-based defense. Note that there is no return value and the adjacency matrix is modified in place.}
\label{fig:similarityCode}
\end{figure}

\paragraph{Modified Nettack code.} Figure~\ref{fig:code} outlines how we changed the Nettack code in order to run the experiments outlined in the paper. 
The function \verb|filter_training| is modeled after \verb|filter_singletons| and is applied analogously immediately afterward. While we pass \verb|trainingData| as a parameter here, this can be computed from the adjacency matrix depending on whether \textsc{StratDegree} or \textsc{GreedyCover} is being used. The amount of training data needs to be specified.

\begin{figure*}[hbt!]
\centering
\footnotesize
\begin{verbatim}
def filter_training(edges, adj, labelVec, trainRatio, trainingData):

    degs = np.squeeze(np.array(np.sum(adj,0)))
    existing_edges = np.squeeze(np.array(adj.tocsr()[tuple(edges.T)]))
    if trainingData is None:  #stratified thresholding based on degree
        nClass = np.max(labelVec)+1
        degThres =  np.zeros((nClass))
        for c in range(nClass):
            temp =  degs[labelVec==c]
            temp = np.sort(temp)
            degThres[c] = temp[int(np.floor(len(temp)*(1-trainRatio)))]

        old_degrees = degs[np.array(edges)]
        if existing_edges.size > 0:
            new_degrees = degs[np.array(edges)] + 2*(1-existing_edges[:,None]) - 1
        else:
            new_degrees = degs[np.array(edges)] + 1

        thresholds = degThres[labelVec[np.array(edges)]]
        breaks = np.logical_or(np.logical_and((old_degrees< thresholds), (new_degrees >= thresholds)), \
                np.logical_and((old_degrees> thresholds), (new_degrees <= thresholds)))
        num_breaks = breaks.sum(1)
        retVal = (num_breaks == 0)
    else:   #use greedyCover
        nonTrain = np.logical_not(trainingData)
        nNonTrainingNeighbors = adj.dot(nonTrain)
        #get the largest number of non-training neighbors among non-training data
        maxNonTrainNeighbors = np.max(nNonTrainingNeighbors[nonTrain]) 
        minNonTrainNeighbors = np.min(nNonTrainingNeighbors[trainingData])
        borderlineTraining = np.logical_and(nNonTrainingNeighbors<= minNonTrainNeighbors+1, trainingData)
        borderlineNonTraining = np.logical_and(nNonTrainingNeighbors >= maxNonTrainNeighbors-1, nonTrain)
        
        
        #don't remove an edge between a borderline  training node and a non-training node,
        #or add an edge betweeen a borderline non-training node and a non-training node
        badRemoval = np.squeeze(np.sum(np.logical_and(borderlineTraining[edges], \
                                                      nonTrain[np.fliplr(edges)]), axis=1)>0)
        badRemoval = np.logical_and(badRemoval, np.logical_not(existing_edges))
        badAddition = np.squeeze(np.sum(np.logical_and(borderlineNonTraining[edges],\
                                                       trainingData[np.fliplr(edges)]), axis=1)>0)
        badAddition = np.logical_and(badAddition, existing_edges)
        
        assert(np.sum(np.logical_and(badRemoval, badAddition))==0)
        breaks = np.logical_or(badAddition, badRemoval)
        retVal = np.logical_not(breaks) #return true if the proposed edge does not break the rule
        
    return retVal
\end{verbatim}
\caption{Modified Nettack code: A function used to remove potential perturbations that would change the composition of the training data.}
\label{fig:code}
\end{figure*}

\end{document}